\documentclass[AMA,Times1COL]{WileyNJDv5}
\usepackage{moreverb}
\usepackage{amsmath}
\usepackage{caption}
\usepackage{subcaption}
\usepackage{multirow}
\usepackage{lmodern}

\usepackage{lineno}
\def\debug{0}

\ifnum\debug=1
\providecommand{\lnote}[1]{\textcolor{blue}{[LR: #1]}}
\providecommand{\tnote}[1]{\textcolor{teal}{[TF: #1]}}
\providecommand{\vnote}[1]{\textcolor{orange}{[VB: #1]}}
\providecommand{\anote}[1]{\textcolor{red}{[AP: #1]}}
\providecommand{\mnote}[1]{\textcolor{violet}{[MB: #1]}}
\renewcommand\hl[1]{%
  \bgroup
  \hskip0pt\color{red!80!black}%
  #1%
  \egroup
}
\else
\providecommand{\lnote}[1]{}
\providecommand{\tnote}[1]{}
\providecommand{\vnote}[1]{}
\providecommand{\anote}[1]{}
\providecommand{\mnote}[1]{}
\renewcommand{\hl}[1]{#1}
\fi

\providecommand{\red}[1]{\textcolor{red}{#1}}

\newcommand\BibTeX{{\rmfamily B\kern-.05em \textsc{i\kern-.025em b}\kern-.08em
T\kern-.1667em\lower.7ex\hbox{E}\kern-.125emX}}

\articletype{Article Type}%

\received{<day> <Month>, <year>}
\revised{<day> <Month>, <year>}
\accepted{<day> <Month>, <year>}

\begin{document}

\title{Swin2-MoSE: A New Single Image Supersolution Model for Remote Sensing}

\author[1]{Leonardo Rossi*}
\author[1]{Vittorio Bernuzzi*}
\author[1]{Tomaso Fontanini}
\author[1]{Massimo Bertozzi}
\author[1]{Andrea Prati}

\authormark{Leonardo Rossi \textsc{et al}}

\address[1]{\orgdiv{Department of Engineering and Architecture}, \orgname{University of Parma}, \orgaddress{\state{Parma}, \country{Italy}} \email{\{name.surname\}@unipr.it}}

\corres{*Leonardo Rossi and Vittorio Bernuzzi, Department of Engineering and Architecture, University of Parma, Parma, Italy. \email{leonardo.rossi@unipr.it} \email{vittorio.bernuzzi@unipr.it}}


\abstract[Abstract]{
Due to the limitations of current optical and sensor technologies and the high cost of updating them, the spectral and spatial resolution of satellites may not always meet desired requirements.
For these reasons, Remote-Sensing Single-Image Super-Resolution (RS-SISR) techniques have gained significant interest.
In this paper, we propose Swin2-MoSE model, an enhanced version of Swin2SR.
Our model introduces MoE-SM, an enhanced Mixture-of-Experts (MoE) to replace the Feed-Forward inside all Transformer block.
MoE-SM is designed with Smart-Merger, and new layer for merging the output of individual experts, and with a new way to split the work between experts, defining a new per-example strategy instead of the commonly used per-token one.
Furthermore, we analyze how positional encodings interact with each other, demonstrating that per-channel bias and per-head bias can positively cooperate.
Finally, we propose to use a combination of Normalized-Cross-Correlation (NCC) and Structural Similarity Index Measure (SSIM) losses, to avoid typical MSE loss limitations.
Experimental results demonstrate that Swin2-MoSE outperforms any Swin derived models by up to $0.377 \sim 0.958$ dB (PSNR) on task of $2\times$, $3\times$ and $4\times$ resolution-upscaling ($\text{Sen2Ven}\mu\text{s}$ and OLI2MSI datasets).
It also outperforms SOTA models by a good margin, proving to be competitive and with excellent potential, especially for complex tasks. Additionally, an analysis of computational costs is also performed.
Finally, we show the efficacy of Swin2-MoSE, applying it to a semantic segmentation task (SeasoNet dataset).
Code and pretrained are available on \url{https://github.com/IMPLabUniPr/swin2-mose/tree/official_code} }


\jnlcitation{\cname{%
\author{Rossi L.},
\author{Bernuzzi V.},
\author{Fontanini T.},
\author{Bertozzi M.}, and
\author{Prati A.}} (\cyear{2024}),
\ctitle{Swin2-MoSE: A New Single Image Super Resolution Model for Remote Sensing}, \cjournal{\lnote{???}}, \cvol{\lnote{???}}.}

\maketitle

\section{Introduction}\label{sec1}

Remote sensing has become more and more important recently for its several applications, especially in the agricultural field.
Crop segmentation and classification, disease detection, automatic irrigation control are some of the possible applications of multi-spectral satellite images that can increase crops productivity and help to fight the waste of resources.
Nevertheless, due to optical and sensor technology, satellites often provide images whose resolution is not enough for their target applications.
For this reason, studying new and powerful remote sensing super resolution techniques is of paramount importance and with this work we were able to provide a novel method that outperforms the current state of the art.

In recent years, the importance of image Super-Resolution (SR), which involves the creation of high-quality images from Low-Resolution (LR) ones, has gained considerable attention from the scientific community~\cite{chen2022real, chauhan2023deep, al2024single}.
The primary goal of SR is to convert blurred, fuzzy, or indistinct LR images with unrefined details into clear and visually appealing high-resolution (HR) images with enhanced details. This augmentation of image details can be beneficial for several image processing and machine learning tasks, including image classification and segmentation.

Considerable research efforts have been dedicated to advancing super-resolution techniques in the more specific field of remote sensing imagery~\cite{liu2021research, fernandez2017single, wang2022review, wang2022comprehensive}.
The literature firstly focused on rectifying the inherently low spatial resolution in images obtained from UAVs or satellites.
Nowadays, with the great availability of satellites that are capable of capturing a multitude of additional spectral bands, such as the near-infrared and infrared spectra, this issue of super-resolution is particularly relevant.
While such multi-spectral imagery offers an increased spectral range of frequencies per-pixel, it concurrently introduces the trade-off of further reduced spatial resolution.
Indeed, the spatial and spectral resolution of Earth Observation satellites typically do not meet the desired high level due to constraints imposed by optical and sensor technology, as well as the associated high costs.
On the other side, many applications could benefit from the availability of higher resolution images, e.g. land use and land cover classification~\cite{kossmann2022seasonet}, object detection~\cite{reiersen2022reforestree}, change detection~\cite{daudt2018urban}, resource management of natural resources as forests, water, and minerals.
For this reason, we decided to focus on the development of a framework which addresses the lack of high resolution multispectral imagery in the context of remote sensing, combining the large availability of low resolution data with a method that allows to reconstruct reliably high frequency details with respect to previous methods.
Moreover, addressing the scarcity of super-resolution methods in remote sensing area, we tailor our framework specifically for satellite imagery which show very different scale and semantics compared to regular RGB imagery on which the vast majority of SR methods are trained for.


SR techniques can be categorized mainly into two types: multi-image super-resolution (MISR) and single-image super-resolution (SISR).
MISR involves the reconstruction of a HR image from a collection of LR images, which are typically obtained from different viewpoints or at different times during satellite passes \cite{anger2020fast, lafenetre2023handheld, nguyen2023l1bsr}.
In this work, the discussion will focus on SISR, a technique in which  the HR image is generated starting from a single LR input.
Compared to MISR, SISR holds particular appeal for several reasons.
Firstly, it entails a reduction in  complexity, as it only needs to process a single image and therefore does not require the alignment and registration of multiple images, which can be computationally expensive.
Secondly, it can be applied to a wider range of scenarios compared to MISR, as it does not require multiple images of the same scene.
Finally, it is easier to implement because it only requires the acquisition of a single image, and less prone to introduce blurring or loss of detail due to the interpolation process.
At the same time, while both MISR and SISR are considered as ill-posed problems because there exist an infinite number of possible HR images that could correspond to a given LR image,
in the case of SISR, the task is more challenging because, by definition, we only have one image available.

In this paper, we propose a novel architecture, called Swin2-MoSE, that exploits Transformer and Mixture-of-Experts blocks together for high-performance SISR. The proposed approach alleviates the complexity associated with designing intricate priors for model-based SISR methods.

The main contributions of this paper are:
\begin{itemize}
    \item A new End-to-End architecture for Single-Image Super Resolution (SISR) called Swin2-MoSE (Swin V2 Mixture of Super-resolution Experts), improving Swin2SR \cite{conde2022swin2sr} baseline.
    \item The introduction of MoE-SM, an enhanced Sparsely-Gated Mixture-of-Experts (MoE) layer \cite{shazeer2017outrageously}, that replaces the Feed-Forward networks inside Transformer building blocks, with the aim of increasing performance.
    It is designed with a new advanced merging layer (Smart Merger, SM) in order to merge the output of individual experts, and with a new way to split the work between experts, defining a new per-example strategy instead of the commonly used per-token one.
    \item An analysis of how positional encodings interact with each other, which led us to integrate a per-channel bias (Locally-enhanced Positional Encoding, LePE \cite{dong2022cswin}) and a per-head bias (Relative Position Encoding, RPE \cite{liu2021swin}).
    \item The proposal of a combination of Normalized-Cross-Correlation (NCC) and Structural Similarity Index Measure (SSIM) losses, in order to avoid typical limitations of the Mean Square Error (MSE) loss.
\end{itemize} \section{Related Works}\label{related}

\noindent\textbf{Single-Image Super-Resolution.}
A super-resolution technique tries to reconstruct a high-resolution image from a low-resolution observation.
Pioneering the resolution enhancement task, SRCNN \cite{dong2015image} employed a three-layer convolutional structure to learn the mapping between LR and HR images.
Subsequent advancements, namely RCAN \cite{dai2019second} and VDSR \cite{kim2016accurate}, addressed the challenges of optimizing deep networks through the incorporation of residual learning techniques.
Despite the lightweight nature of these approaches, the convolution kernel usually has a limited receptive field and hinders the extraction of long-range or non-local features.
As a result, for some regions with fine	details, these methods exhibit sub-optimal performance.
With the emergence of the Generative Adversarial Network (GAN) framework, SRGAN  \cite{ledig2017photo} was introduced showcasing its ability in super-resolution tasks by generating photo-realistic images, even for large upscaling factors, leveraging content, perceptual, and adversarial losses.
ESRGAN \cite{wang2018esrgan} refines the latter method by removing Batch Normalization layer, introducing Residual-in-Residual Dense Blocks, and enhancing the perceptual loss through Mean Squared Error (MSE) calculations on high-level feature maps before activation.
For capturing long-distance spatial contextual information Second-order Attention Network (SAN) \cite{dai2019second} introduces a non-locally enhanced residual group and implements a novel second-order channel attention module to learn feature inter-dependencies by considering second-order statistics of features.
Holistic Attention Network (HAN) \cite{niu2020single} improves inter-layer dependencies through a Layer Attention Module and augments channel and spatial inter-dependencies using a Channel-Spatial Attention Module, addressing some issues from previous method.
The UFormer \cite{wang2022uformer} introduces U-shaped encoder-decoder structure with a window-based system to alleviate self-attention calculation, similar to the Swin Transformer \cite{liu2021swin} approach.
Building upon the Swin transformer architecture, authors of SwinIR \cite{liang2021swinir} introduced a novel Residual Swin Transformer Block for deep feature extraction exhibiting noteworthy performance for this task.
In the SwinFIR \cite{zhang2022swinfir}, authors replaced the convolutions ($3 \times 3$) inside the SwinIR with a Fast Fourier Convolution component to have the image-wide receptive field.
Innovating upon SwinIR, Swin2SR \cite{conde2022swin2sr} incorporates the Swin Transformer V2 \cite{liu2022swin} block, enhancing scalability for higher resolution input images, ensuring stability during training, and improving overall efficiency.
All these models, descending from a Swin architecture, contain a small MLP within the Transformer block which may not be able to adapt effectively to the growth of available data.
Inspired by the successes of MoE \cite{riquelme2021scaling, jordan1994hierarchical, shazeer2017outrageously} architectures, we have successfully explored its use in super-resolution, introducing an enhanced Sparse Mixture of Experts building block, called MoE-SM, instead of MLP inside the Transformer block.
Authors of DAT \cite{conf/iccv/0014ZGKY023} proposed to use spatial and channel self-attention together to exploit both context information.
Like them, we too felt the need for spatial and channel information.
In our case however, we use two different positional encodings within the same self-attention.
In addition, unlike the others, in our work we explore the use of a combination of losses, to overcome the limits related to overfit on low frequencies of the signal.

\noindent\textbf{Remote Sensing Single-Image Super-Resolution.}
In prior investigations, conventional super-resolution techniques like SRCNN \cite{tuna2018srcnnrs} and VDSR \cite{dong2016vdsrrs}, employing IHS (Intensity, Hue, Saturation) transformations, have demonstrated promising performance when applied to very high resolution satellite data obtained through SPOT6/7 and Pleiades 1A/1B platforms. However, due to the intricate nature of remote sensing imagery, it becomes imperative to design specialized models tailored specifically for Remote Sensing Single-Image Super-Resolution (RS-SISR) \cite{karwowska2022review}.
To address this requirement, Lei et al. \cite{lei2017lgcnet} developed Local-Global Combined Network, featuring a local-global fusion architecture that leverages multiple levels of representation learned via concatenation among various convolutional layers within the network. Xu et al. \cite{xu2018dmcn} presented Deep Memory Connected Network, which integrates both image details and environmental context into its framework, utilizing local and global memory connections. Ren et al. \cite{ren2021ercnn} introduced Enhanced Residual Convolutional Neural Network with Dual-Luminance Scheme, enhancing the feature flow module's capability while improving the distinction between features across different map scales.
Additionally, Generative Adversarial Networks (GANs) have been employed in real-world RS-SISR applications, as exemplified by Guo et al.'s \cite{guo2022ndsrgan} introduction of Novel Dense GAN alongside their Real High-Resolution and Low-Resolution Aerial Imagery dataset, consisting of paired high-resolution and low-resolution aerial photographs captured under varying elevations. Furthermore, Zhang et al. \cite{zhang2022rbanunet} introduced an integrated framework to simulate remote sensing image degradation, encompassing blur kernel estimation and noise pattern extraction.  They utilize a Residual Balanced Attention Network along with a modified U-NET model acting as a discriminator, thereby boosting perceptual performance.
Xiao et al. \cite{xiao2023transformerRS} devised a Cross-scale Hierarchical Transformer (CHT) that excels in modeling both global and local cross-scale representations. CHT employs two distinct modules — cross-scale self-attention and cross-scale channel attention — to discern meaningful patterns across varied spatial scales.
Tu et al. \cite{tu2022swcgan} suggested a Generative Adversarial Network incorporating a fusion of convolutional and Swin Transformer layers for super-resolution reconstruction of remote sensing images. Their method emphasizes large-sized, informative, and correlated pixel properties specific to remote sensing imagery.

\noindent\textbf{Positional Encoding.}
In Transformer models, since self-attention is invariant to token permutations and does not consider positional information of tokens, positional encodings are commonly utilized to reintroduce this lost positional data.
Several approaches have been designed, including Absolute Positional Encoding (APE) \cite{vaswani2017attention}, Relative Positional Encoding (RPE) \cite{shaw2018self,liu2021swin} and Locally-Enhanced Positional Encoding (LePE) \cite{dong2022cswin}.
The first one is also the most used and consists in a sinusoidal function added to the inputs to inject the absolute position of the token.
The second encoding (RPE) focuses on capturing the pairwise relationships between elements in the input sequence.
This method is particularly useful for tasks where the relative ordering or distance between elements is important and adds the translation equivariance property, similar to convolutions.
The last one (LePE) introduces a per-channel bias, rather than the RPE which implement a per-head bias.
Similarly to RPE, in Swin Transformer v2 \cite{liu2022swin, conde2022swin2sr}, a log-spaced Continuous Position Bias (log-spaced CPB or log CPB) approach is introduced to improve extrapolation accuracy when transferring between large size changes across windows sizes, by compressing large value ranges into a smaller, more manageable scale.
For this purpose, they adopted a Continuous Relative Position bias, integrating a small MLP for translating relative coordinates.
Since RPE and LePE possess interesting and useful characteristics for the super-resolution task, unlike what others do, both were used in our model, obtaining both relative position information and improved locality information.

\noindent\textbf{Mixture-of-Experts (MoE).}
The Mixture-of-Experts approach was introduced in natural language processing (NLP) within a recurrent language model \cite{shazeer2017outrageously}. It was used to implement conditional computation, allowing the model to scale up to billions of weights, while limiting the increase in computational cost.
Sparse-Mixture-of-Experts models \cite{jordan1994hierarchical} are a type of neural network architecture that uses a subset of specialized sub-networks (experts) to improve performance on a given task.
The sparsity \cite{shazeer2017outrageously}, obtained by training a gate network, permitted to increase the model size significantly, without sacrificing efficiency too much.
In a related study \cite{10.1007/978-3-031-19797-0_33}, authors use a small regression network for image degradation modelling.
Simultaneously, multiple convolutional experts are jointly trained within the MoE framework to collectively cover the degradation space.
During evaluation, adaptive combination of expert parameters yields a singular optimized network for inference, by that enhancing efficiency.
In the paper \cite{he2024frequency}, authors use specialized low-frequency and high-frequency MoE blocks together to ensure the efficient reconstruction of both low-frequency and high-frequency information.
In the paper \cite{hwang2023tutel}, authors introduced MoE inside a Swin Transformer model for the classification task, by replacing the MLP, except for the first two network stage.
Similarly to Switch Transformer \cite{fedus2022switch}, SwinV2-MoE \cite{hwang2023tutel} and V-MoE \cite{riquelme2021scaling}, we decided to replace the MLP of the Transformer block with a MoE.
Differently from the first, we made the design choice to use more than one expert each time, with the resulting routing algorithm.
While in SwinV2-MoE the first two network stages were not replaced, in our model all the MLP layers are substituted with MoE.
Due to spatial proximity in images, it is very likely that nearby tokens are similar, in our opinion it is much more favorable that all the tokens of an example are processed by the same experts.
or this reason, differently from the others, we have chosen a per-example strategy instead of a per-token one to feed the experts.
This should facilitate the specialization ability of each expert and reduce the fragmentation of experts in the case of distributed training.
Furthermore, one of the aspects less explored by other methods is linked to the fusion of the results of individual experts.
For this reason, we designed a new way to merge experts output with a Smart Merger (SM) layer.

 \section{The Proposed Swin2-MoSE Model}
\label{theory}

\subsection{Overall Network Architecture}

\begin{figure}[ht!]
\centerline{\includegraphics[clip, trim=0 0 0 0, width=.8\linewidth]{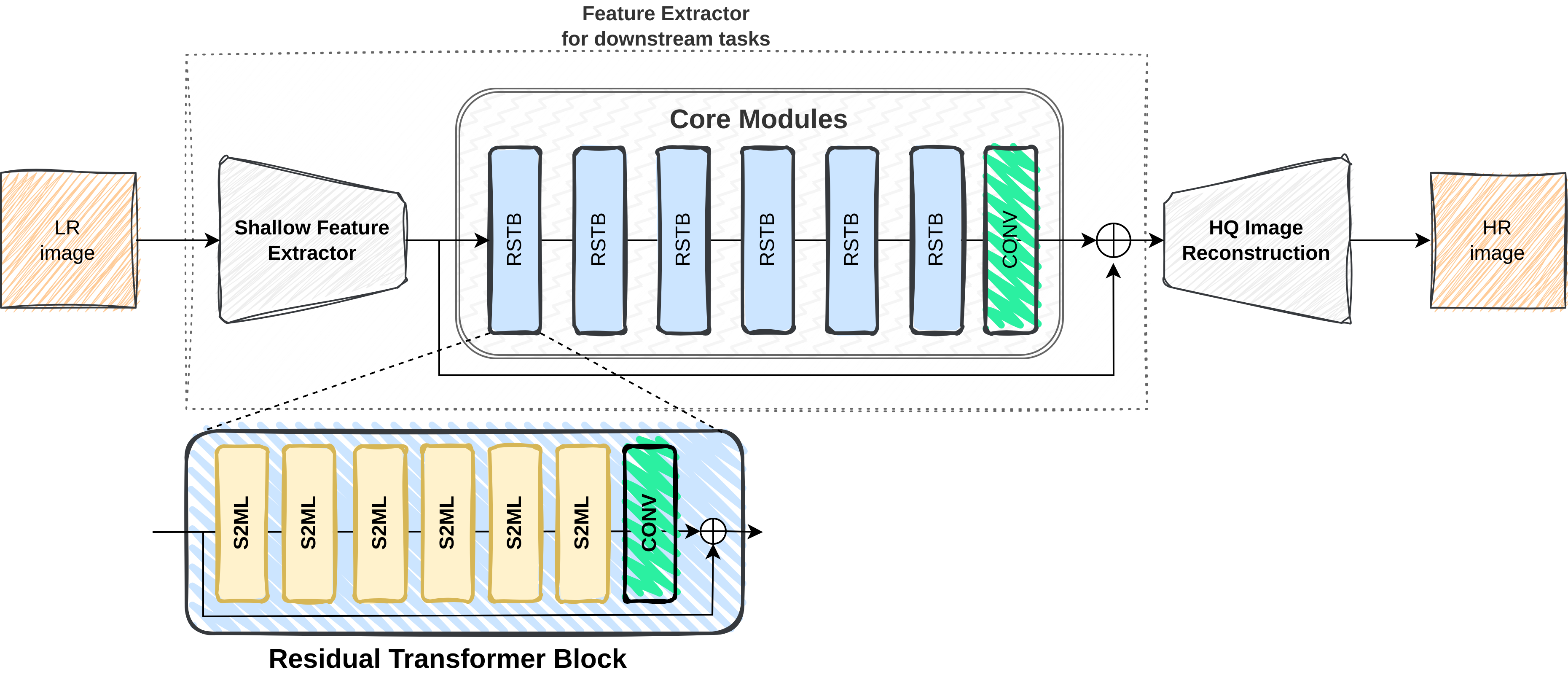}}
    \caption{The architecture of the proposed Swin2-MoSE (solid colors for the architectural changes introduced). Note: the Feature Extractor will be discussed in experiments section \ref{sec4.5} related to the semantic segmentation task.}
    \label{swin2-moe-arch}
\end{figure}

Figure \ref{swin2-moe-arch} shows the overall design of the Swin2-MoSE system.
Based on the limitations of the original Swin2SR \cite{conde2022swin2sr} model, the following subsections will describe the changes proposed in our model to improve the ability to perform super-resolution tasks.
In subsection~\ref{moe-sm theory}, we propose the introduction of a new MoE module (called MoE-SM) inside the Transformer blocks (called S2ML, Swin v2 MoSE Layer).
Then, in subsection~\ref{positiona_enc}, how positional encodings interact with each other will be discussed, and some conclusions about their use on Transformer architecture for the super-resolution task will be drawn.
Finally, in subsection~\ref{losses}, we also investigate various loss functions to improve the robustness and performance of the model training.

\subsection{Mixture of Experts}
\label{moe-sm theory}

\begin{figure}[ht!]  \centerline{\includegraphics[clip, trim=0 0 0 0, width=.8\linewidth]{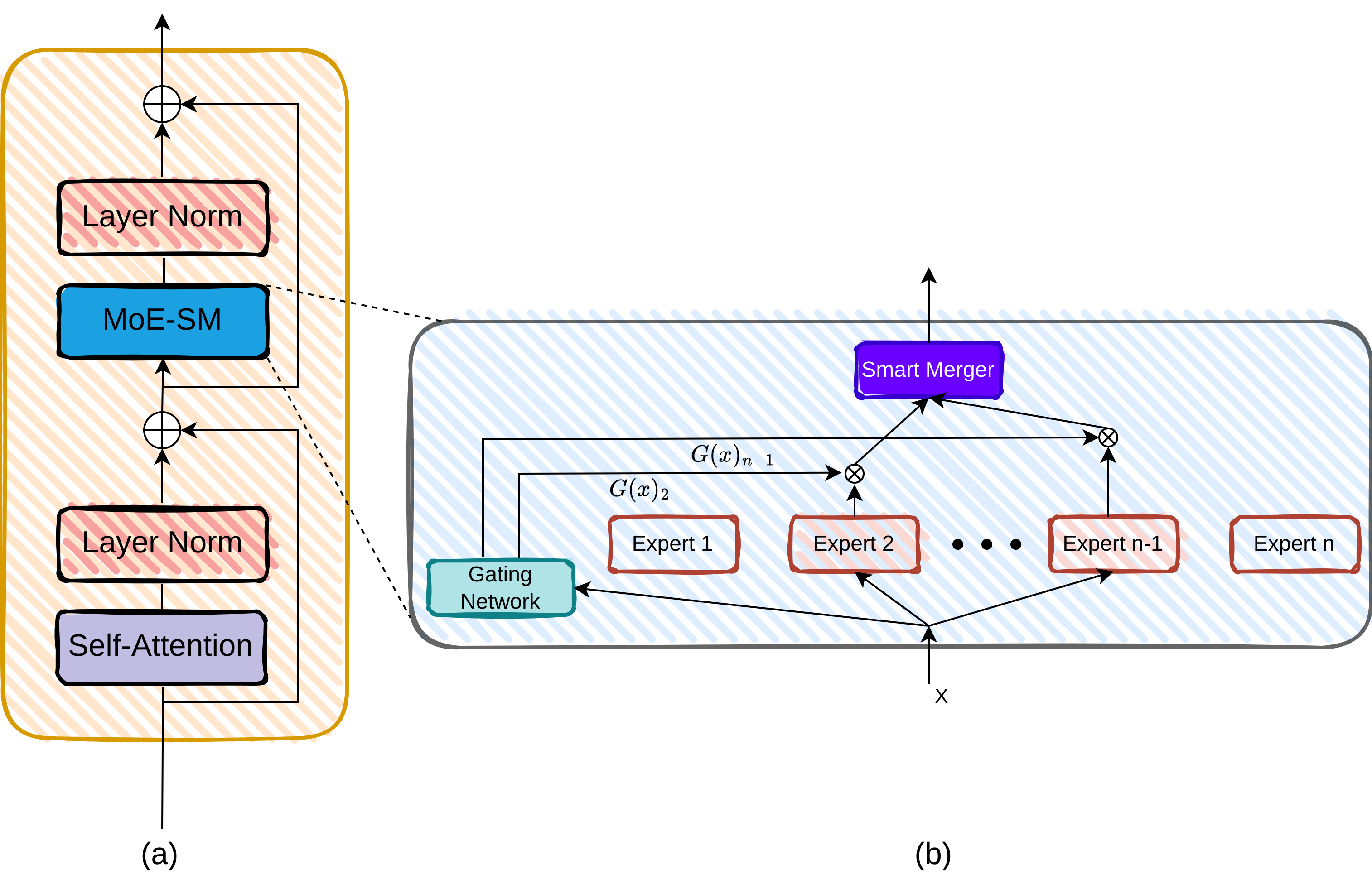}}
    \caption{Mixture of Experts architecture (solid colors for the architectural changes introduced): (a) Our Swin v2 MoSE Layer (S2ML) with a Self-Attention block (see Figure \ref{fig:pos-enc} for more details) and a MoE-SM in the place of MLP. (b) MoE-SM internal architecture with our Smart Merger module to merge experts outputs.}
    \label{moe-arch}
\end{figure}

Conditional computation paradigms have become popular in deep learning because they allow to scale model capacity, particularly in response to large datasets, maintaining a fixed compute budget.
Mixture of Experts (MoE) architecture uses this approach to implement sparsity, improving efficiency and performance by only exploiting necessary parts of the model for a given input.
Here, the degree of sparsity defines the ratio between the number of blocks that are actually used to compute the output, compared to the total available.
In practical terms, the concepts of ``sparse parameter count'' (SPC) and ``active parameter count'' (APC) ~\cite{jiang2024mixtral} are decoupled, where the first indicates the model total parameter count, and the second indicates the number of parameters used for processing an individual input.
This has been achieved by allowing models to selectively activate only the parts of the network that are relevant to the task at hand, rather than activating the entire network for every input.
A comprehensive investigation was done firstly in the domain of NLP, scaling a LSTM model up to 137B parameters \cite{shazeer2017outrageously}.
The concept of conditional computation has been even more explored in recent years, culminating in the development of models with a staggering 1.6T parameters \cite{fedus2022switch}.

The MoE architecture is composed by a number of Experts that are individually able to elaborate the input, a Gate Network (GN) to actively select a subset of the Experts and a final merging layer (Smart Merger) to fuse the outputs of individual Experts and produce the final result.
The MoE should allow the experts to focus on different domains, resulting in a more diverse and accurate management of complex problems.
Our objective here is to judiciously leverage this capability of the MoE architecture in a super-resolution context, with the aim of optimizing the performance of our model, improving quality of its results, by exploiting the experts specialization.
Simultaneously, this choice would prepare for a future model scaling, increasing the number of experts, without incurring in an unbearable computational cost.

Our model uses as building block the Swin2SR \cite{conde2022swin2sr} Transformer block, which represents the state-of-the-art model in the super-resolution field.
As shown in Figure \ref{moe-arch}(a), similarly to Switch Transformer \cite{fedus2022switch}, SwinV2-MoE \cite{hwang2023tutel} and V-MoE \cite{riquelme2021scaling}, we decided to replace the MLP of the Transformer block with our MoE-SM module.
We replaced all the MLP layers in the blocks, because we believe that MoE-SM modules can significantly contribute to all processing levels, without adding excessive overhead.

Our MoE-SM (see Figure \ref{moe-arch}(b)) is designed as a Sparsely-Gated Mixture-of-Experts \cite{shazeer2017outrageously} but, differently from the original version, it is designed with an advanced merging layer (Smart Merger, SM) in order to merge the output of individual experts.
Our SM is composed by a 2D convolutional layer with the number of input channels equals to the number of active experts per batch element $k$ and output channels of 1 (with a kernel size of 3, padding 1).
In this way, we implement an efficient and learnable weighted sum of the input as opposed to a simple and not learnable weighted sum, employed in previous works.

Finally, a trainable Gating Network is used to select the most appropriate experts.
Additionally, our MoE-SM works with a new per-example strategy instead of a usually used per-token one.
More in detail, the GN redirects each input example in the batch to a subset of experts, rather than distributing individual tokens.
Unlike the authors of the paper \cite{shazeer2017outrageously}, we do not use the noise term within the gating network, but, instead, we use a more simple Top-K Gating mechanism.
We chose this solution because we noted that the learned noise continues to grow during the training, destabilizing the learning process of the Gating Network.
The formula for the Top-K Gating Network is presented below.

\begin{equation}
\begin{split}
G \left(x \right) &= \text{Softmax} \left( \text{KeepTopK} \left( x \cdot W_g\right),  k \right) \\
\text{KeepTopK} \left(v, k \right) &= \begin{cases}
   v_i \qquad \text{if} \ v_i \ \text{is in the top} \ k \ \text{elements of} \ v \ \\
   -\infty \quad \text{otherwise}
\end{cases}
\end{split}
\label{eq:topk}
\end{equation}

\noindent where $G\left(x\right)$ are the softmaxed Top-K gating values, $W_g$ are the learnable weights of the Gating Network, $x$ are the input features and $k$ is the number of experts active for each time example.

As in the original implementation \cite{shazeer2017outrageously}, we use a loss inside the MoE-SM to promote expert balancing utilization.
A metric called $Imp(X)$ is defined to evaluate the relative significance of an expert within a given set of examples.

\begin{equation}
\begin{split}
Imp(X) &= \sum_{x \in X} G \left (x \right ) \\
\mathcal{L}_{MoE} &= CV\left(Imp\left(X\right)\right)^2 \\
\end{split}
\label{eq:loss-importance}
\end{equation}

\noindent The $CV$ is the coefficient of variation of a sample, defined as the ratio between the variance and the squared mean of $X$.
The $\mathcal{L}_{MoE}$ is the total MoE-SM loss.
 \subsection{Positional Encoding}
\label{positiona_enc}

\begin{figure}[ht]
    \centerline{\includegraphics[clip, trim=0 0 0 0, width=.8\linewidth]{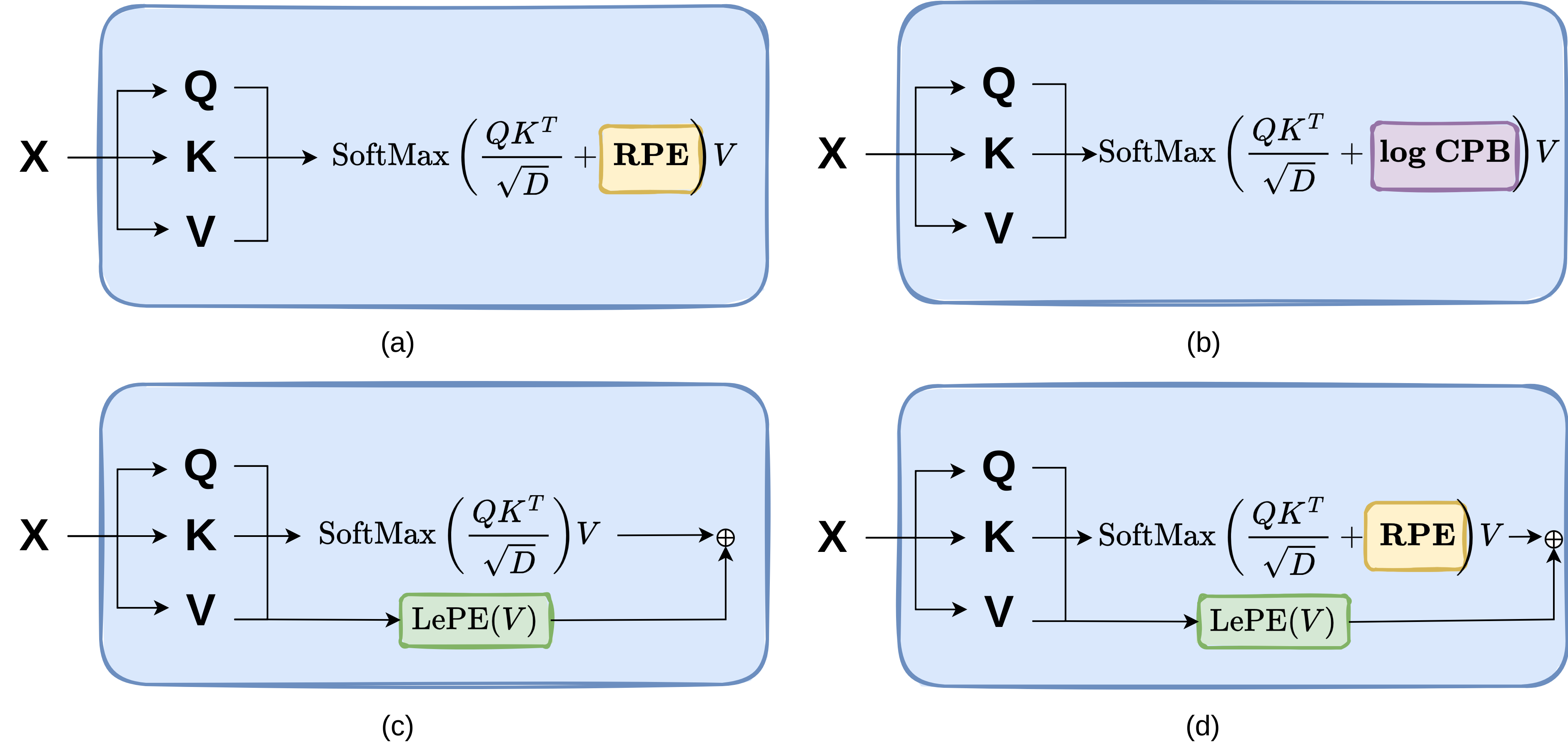}}
    \caption{Comparison between positional encoding methods inside the self-attention block:
    (a) Relative Position Encoding (RPE) proposed on Swin Transformer;
    (b) Log-Spaced CPB (log CPB) proposed on Swin Transformer v2;
    (c) Locally-enhanced Positional Encoding (LePE) proposed on CSwin Transformer;
    (d) Our solution that integrates multiple positional encoding.}
    \label{fig:pos-enc}
\end{figure}

The self-attention operation in Vision Transformers is permutation-invariant, leading to the loss of important positional information in 2D images.
To address this issue, various positional encoding mechanisms have been incorporated to preserve spatial relationships between image regions.
Indeed, the most promising ones are the methods that introduce per-head bias, such as RPE \cite{liu2021swin} and log-spaced CPB \cite{liu2022swin}, and those which introduce a per-channel bias, such as LePE \cite{dong2022cswin}.
To the best of our knowledge, apart for APE and RPE inside the Swin Transformer \cite{liu2021swin}, there has been no prior attempt to analyze how the most popular positional encodings interact with each other when used together.

Our experiments revealed at least two significant findings:
\textit{(i)} the combination of per-head bias and per-channel bias yields better results than either alone, as they contribute distinct positional information to the final output -- we can say that the first provides a two-dimensional positional encoding on width and height, while the second injects a one-dimensional positional encoding on a third dimension, represented by the channel; \textit{(ii)} using two per-head bias together typically does not bring substantial improvements to justify the computational overhead, but it does not conflict each other, therefore, one per-head bias is usually sufficient.

As shown in Figure \ref{fig:pos-enc}(d), we selected the combination of RPE and LePE because it provides the highest benefits to the super-resolution task, as it will be shown in the experimental results. This combination, indeed, offers both advantages: relative position information and improved locality.
 \subsection{Losses}
\label{losses}

Two of the most used metrics to evaluate the super-resolved image are Peak Signal-to-Noise Ratio (PSNR) and Structural Similarity Index Measure (SSIM) metrics.
However, PSNR does not always provide the correct value between overly smoothed image and a sharp one\cite{sajjadi2017enhancenet}, and its values can vary significantly even for nearly indistinguishable images.
Similarly, while SSIM is capable of evaluating brightness, contrast, and structure, it still falls short in reliably quantifying perceptual quality \cite{ledig2017photo}.
Whenever these metrics are employed as loss functions during the training process, it is evident that their limitations will be reflected in the super-resolved images generated by the model.
In other words, the weaknesses of these metrics will be propagated to the output images, potentially resulting in sub-optimal performance.

In order to minimize the distortions present in the high-resolution versions, we conducted experiments to evaluate the impact of various reconstruction metrics when used as losses. In an attempt to find a more appropriate loss for this task, ultimately, we concluded that the combination of Normalized-Cross-Correlation (NCC) loss and Structural Similarity Index Measure (SSIM) loss produced the most favorable results, effectively limiting the number of distortions introduced during the super-resolution process.
Although both NCC and SSIM losses are considered similarity measures, they indeed rely on global and local statistics, respectively.
In our opinion, this feature allowed the model to obtain better quality results.

The overall loss is described below as the sum of the following weighted components:

\begin{equation}
\mathcal{L} = \alpha \cdot \mathcal{L}_{NCC} + \beta \cdot \mathcal{L}_{SSIM} + \gamma \cdot \mathcal{L}_{MoE}
\end{equation}

\noindent where $\alpha$, $\beta$ and $\gamma$ are the weights of NCC, SSIM and MoE losses, respectively.

\noindent\textbf{NCC Loss.}
The Normalized-Cross-Correlation (NCC) or Pearson correlation coefficient metric score is a measure of the similarity between two signals or images, and it is calculated by taking the inner-product between two different signals/images and their mean value.
The score ranges from -1 to 1, where a score of 1 indicates perfect correlation, a score of 0 indicates no correlation, and a score of -1 indicates perfect anti-correlation.

The equation for calculating the NCC value for the $s$-th spectral channel ($NCC_s$) is the following:

\begin{equation}
NCC_s = \frac{\sum \left(\hat{y} - \mu_{\hat{y}}\right) \cdot \left(y - \mu_{y}\right)}{\sqrt{\sum \left(\hat{y} - \mu_{\hat{y}}\right)^2 \cdot \sum_{i=1}^{n} \left(y - \mu_{y}\right)^2}}
\end{equation}

\noindent where $\hat{y}$, $y$ are the output of the network and the ground-truth, respectively, and
$\mu_{\hat{y}}$, $\mu_{y}$ are the mean of all $\hat{y}$ and $y$ values.
Each channel is treated separately, obtaining a value of NCC per channel.
Finally, the average across all spectral channels is returned.
The closer the NCC values are to 1, the better the network is performing.
Since the NCC value can take on negative values, the corresponding loss function is defined below, modifying the possible ranges of values:

\begin{equation}
\begin{split}
NCC &= \frac{1}{S} \sum_{s=1}^{S} NCC_s \in [-1, 1]\\
\mathcal{L}_{NCC} &= 1 - \frac{1}{2}\left(NCC + 1\right) \in [0, 1]
\end{split}
\end{equation}

\noindent where $\mathcal{L}_{NCC}$ is our NCC loss, $S$ is the number of spectral channels.

\noindent\textbf{SSIM Loss.}
Structural Similarity Index (SSIM) \cite{wang2004image} is a metric used to measure the similarity between two images too.
The score is calculated by comparing the luminance, contrast, and structural information of two images.
The equation for calculating the SSIM value for the s-th spectral channel ($SSIM_s$) is the following:

\begin{equation}
\begin{split}
SSIM_s\left(\hat{y},y\right) &= \mathbb{L}\left(\hat{y},y\right)^\delta \cdot \mathbb{C}\left(\hat{y},y\right)^\varepsilon \cdot \mathbb{S}\left(\hat{y},y\right)^\eta\\
\mathbb{L}\left(\hat{y},y\right) &= \frac{2\mu_{\hat{y}}\mu_y + C_1}{\mu_{\hat{y}}^2 + \mu_y^2 + C_1}\\
\mathbb{C}\left(\hat{y},y\right) &= \frac{2\sigma_{{\hat{y}}y} + C_2}{\sigma_{\hat{y}}^2 + \sigma_y^2 + C_2}\\
\mathbb{S}\left(\hat{y},y\right) &= \frac{\sigma_{{\hat{y}}y} + C_3}{\sigma_{\hat{y}}\sigma_y + C_3}
\end{split}
\end{equation}

\noindent where $\sigma _{\hat{y}}^{2}$, $\sigma _{y}^{2}$ and $\sigma _{\hat{y}y}$ are the variance and covariance values of $\hat{y}$ and $y$, respectively, and
$C_{n}$ with $n\in[1, 3]$ are small constants to avoid instability.
A $11\times11$ circular-symmetric Gaussian Weighing function is applied pixel-by-pixel over the entire image.
At each step, the local statistics and $SSIM_s$ index are calculated within the local window, which moves pixel-by-pixel over the entire image.
The luminance measure $\mathbb{L}\left(x,y\right)$ compares the mean intensity values of the two images, while the contrast measure $\mathbb{C}\left(x,y\right)$ compares the standard deviations of the two images, and the structural similarity measure $\mathbb{S}\left(x,y\right)$ compares the correlation between the two images.
$\delta$, $\varepsilon$, and $\eta$ are parameters that control the relative importance of luminance, contrast, and structural similarity, respectively.
Typically, $\delta=\varepsilon=\eta=1$.

Finally, the SSIM loss, called $\mathcal{L}_{SSIM}$, is defined as the average of SSIM across all $S$ spectral channels:

\begin{equation}
\begin{split}
\mathcal{L}_{SSIM}\left(\hat{y}, y\right) = \frac{1}{S} \sum_{s=1}^{S} \left(1 - SSIM_s\right)
\end{split}
\end{equation}
  \section{Experimental Results}\label{experiments}



\noindent\textbf{Tasks and Datasets.}
We performed several experiments on different datasets.
In particular, for the ablations, we tested the $4\times$ scale on $\text{Sen2Ven}\mu\text{s}$ dataset.
To directly assess the performance of SISR models, we used both $\text{Sen2Ven}\mu\text{s}$ \cite{michel2022sen2venmus} and OLI2MSI \cite{wang2021multisensor}.
Both datasets consist of pairs of LR and HR images obtained from different satellites, each with distinct resolutions.
Finally, we assess the indirect impact of super-resolution image quality in a practical application, namely semantic segmentation, using the SeasoNet \cite{kossmann2022seasonet} dataset.
Below, a more detailed description of the chosen datasets can be found.

\noindent\textbf{$\text{Sen2Ven}\mu\text{s}$ dataset.}
This is a large-scale Single Image Super-Resolution dataset, made publicly available for the remote sensing research community.
It comprises 10~m and 20~m ($128\times128$ and $64\times64$ pixels) patches from Sentinel-2, accompanied by their corresponding 5~m ($256\times256$ pixels) resolution patches, which were acquired by the $\text{Ven}\mu\text{s}$ satellite on the same day and within a maximum time difference of 30 minutes.
It encompasses 29 locations captured across 2 years, for a total of 132,955 patches.

\noindent\textbf{OLI2MSI dataset.}
The dataset is also a collection of paired real-world multi-sensor LR-HR data, by carefully selecting relatively cloud-free images in the same location, obtained within a suitable temporal window.
The chosen satellites are Landsat-8 OLI, with a 30~m resolution patches ($160\times160$ pixels) for LR, and Sentinel-2 MSI, with a 10~m resolution patches ($480\times480$ pixels) for HR.
The dataset is composed by 5325 pairs, randomly divided into two parts: 5225 pairs for training set and 100 pairs for testing set.

\noindent\textbf{SeasoNet dataset.}
This extensive multi-label land cover and land use comprehension dataset comprises 1,759,830 scenes derived from Sentinel-2 satellite tiles, featuring 12 spectral bands.
Each individual patch attains dimensions of $120\times120$ pixels, $60\times60$ pixels, $20\times20$ pixels, corresponding to the 10~m, 20~m, and 60~m Sentinel-2 bands, respectively.
SeasoNet pairs each Sentinel-2 image with a segmentation map with a shape of $33 \times 120 \times 120$, where 33 are the defined classes in SeasoNet.
A total of 519,547 unique patch locations were sampled across the dataset.
For each designated patch location, the collection of images was stratified according to seasons, resulting in five separate datasets, along with an additional snowy subset.
To assess the super-resolved quality of the images, the subset representative of the Spring season, grid 1, was selected for training and evaluation.
For each image, the training process employed the 10m bands, namely B2, B3, B4, and B8.

\noindent\textbf{Implementation details.}
Unless otherwise specified, for all experiments, we perform the training on a single A100 GPU, with Adam optimization algorithm, learning rate of 0.0001, 70 epochs, batch size 8.
In the case of last experiment on SeasoNet, we trained with Adam optimization algorithm, learning rate of 0.0001, 40 epochs, batch size 128.

\noindent\textbf{Evaluation Metrics.}
In order to make quantitative comparisons, we take into account two metrics: the Peak Signal-to-Noise Ratio (PSNR), measured in decibel (dB), and the Structure Similarity Index Measure (SSIM).
Only in the first ablation about the losses in section~\ref{experiment_losses}, for reasons related to the experiment itself, we employ also a Normalized-Cross-Correlation (NCC) metric.
In the case of last experiment on SeasoNet, F1-macro and F1-micro metrics are used to evaluate the performance.
F1-macro calculates the average F1 score for each class and treats them equally, while F1-micro takes into account the number of samples in each class and gives more weight to correct predictions for the majority class.
For all of them, the higher the better.

\subsection{Training Losses}
\label{experiment_losses}


\begin{figure}[ht!]
    \centering
    \begin{subfigure}{0.33\linewidth}
        \includegraphics[width=1.0\linewidth]{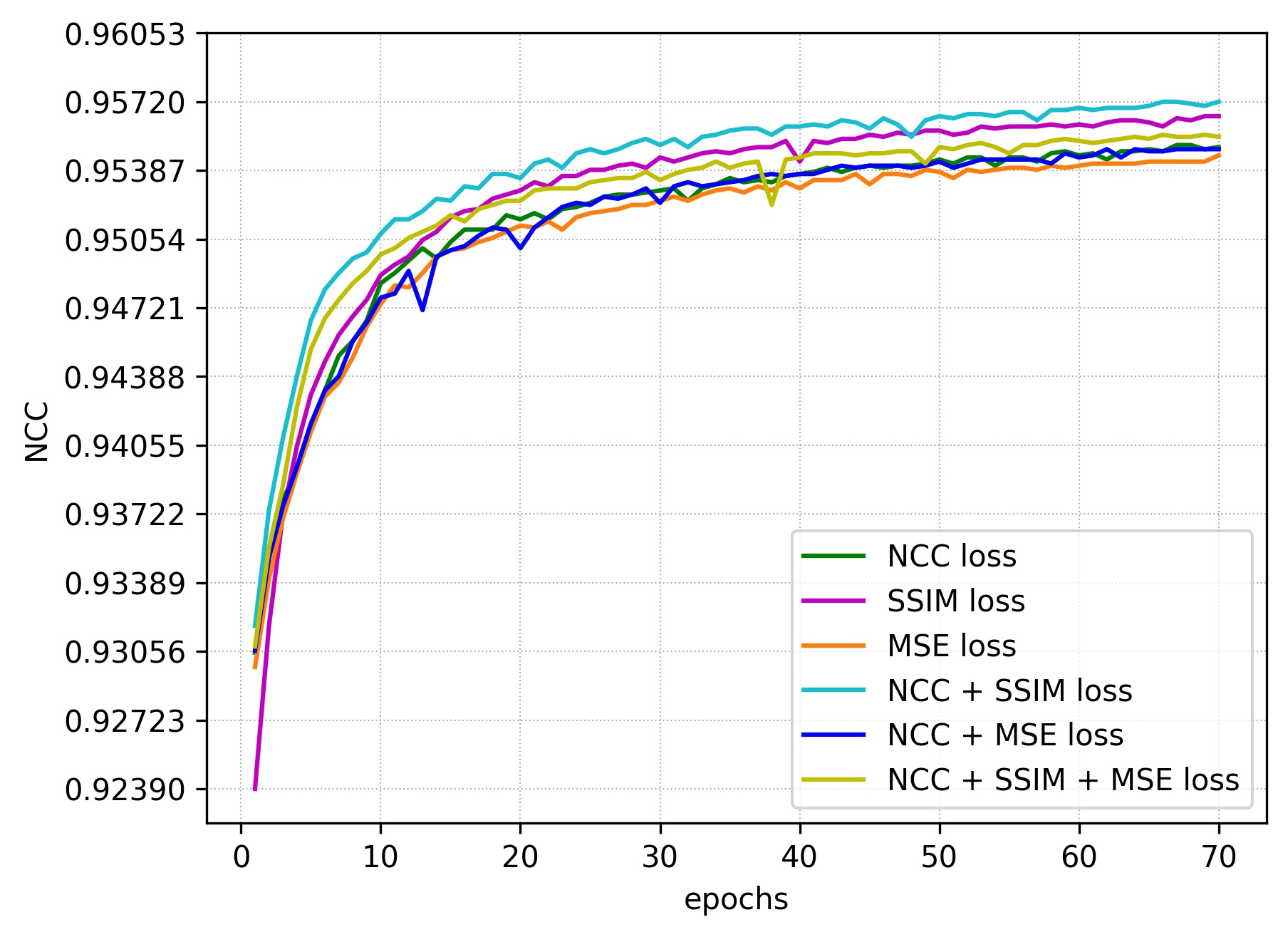}
        \label{fig:loss-cc}
    \end{subfigure}
    \begin{subfigure}{0.33\linewidth}
        \includegraphics[width=1.0\linewidth]{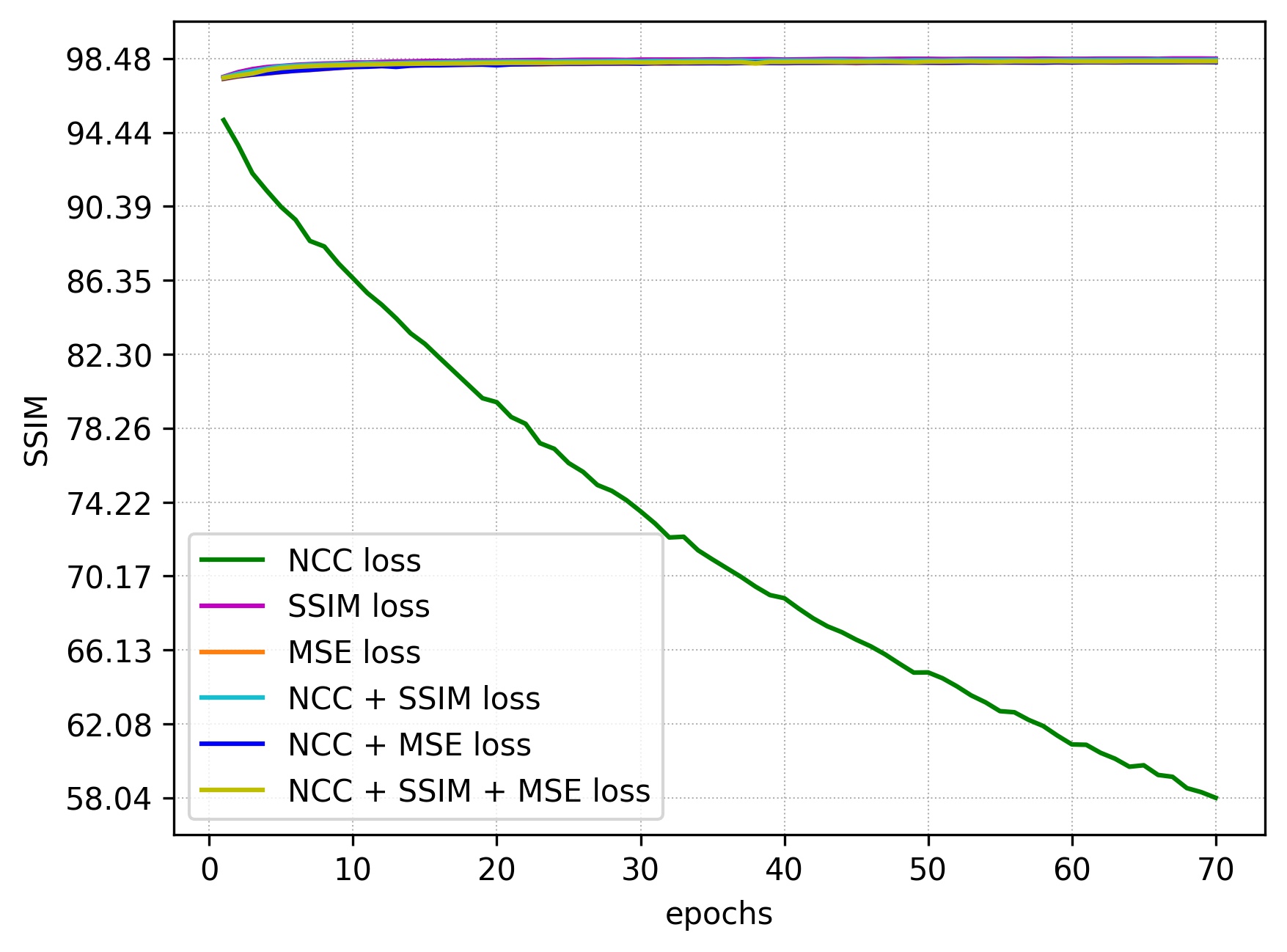}
        \label{fig:loss-ssim}
    \end{subfigure}
    \begin{subfigure}{0.33\linewidth}
        \includegraphics[width=1.0\linewidth]{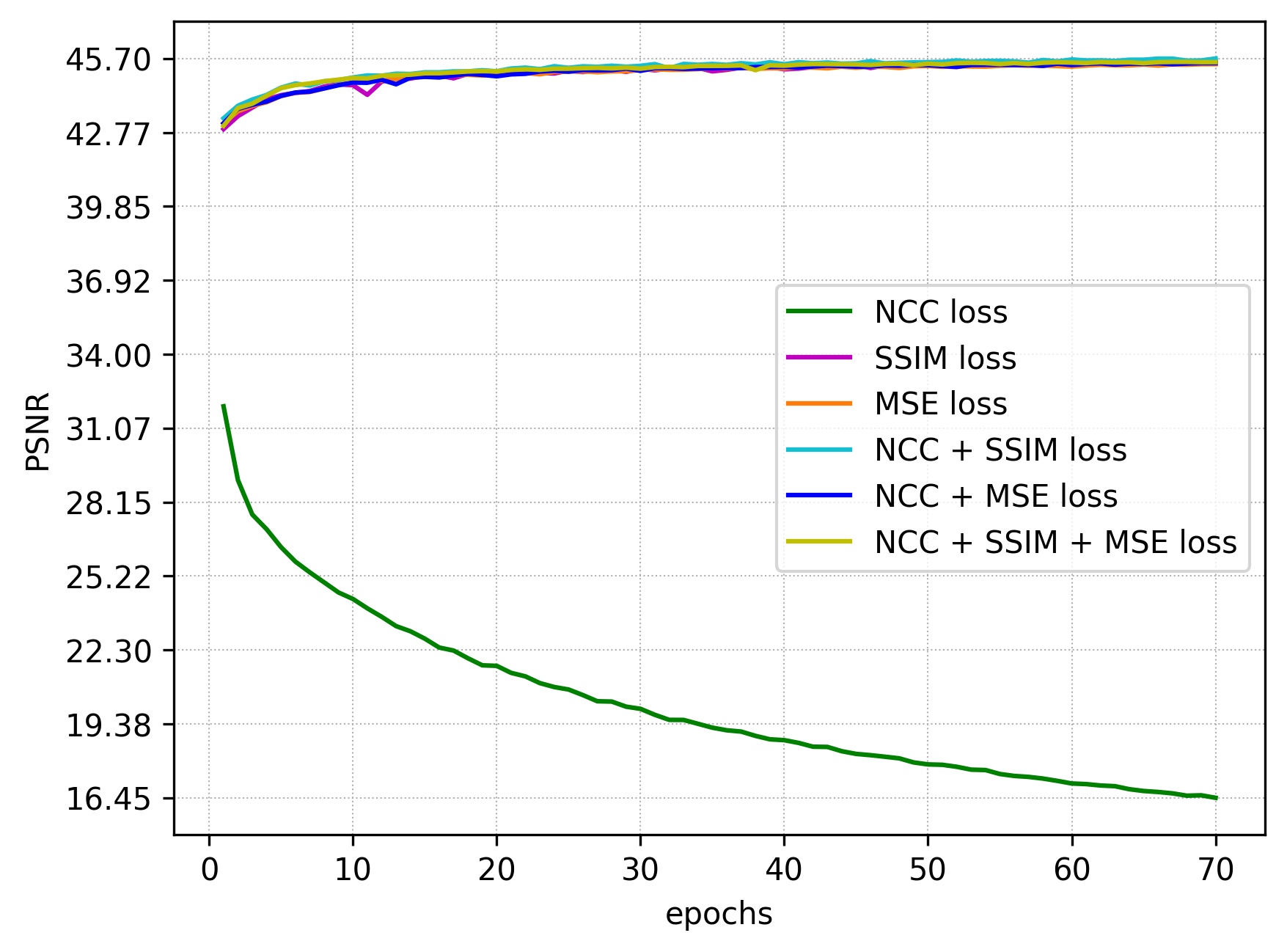}
        \label{fig:loss-psnr}
    \end{subfigure}
 \caption{Trends of the NCC, SSIM, and PSNR metrics for different loss combinations.}
\label{fig:losses-comparison}
\end{figure}

\begin{table}[ht!]
\caption{Ablation study on loss usage. The table shows the performance of the model using different combinations of loss functions. The combination with the highest performance is shown in \textbf{bold} and the second-highest performance is shown in \textcolor{red}{red}.\label{tab:loss-comparison}}
\centerline{\begin{tabular}{|c|c|c|c|l|l|l|}
\hline
  & \multicolumn{3}{c|}{Losses} & \multicolumn{3}{c|}{Performance} \\
  & NCC & SSIM & MSE & NCC $\uparrow$ & SSIM $\uparrow$ & PSNR $\uparrow$ \\ \hline
1 & \checkmark &            &            & 0.9550 & 0.5804 & 16.4503 \\
2 &            & \checkmark &            & \red{0.9565} & \textbf{0.9847} & 45.5427 \\
3 &            &            & \checkmark & 0.9546 & 0.9828 & 45.4759 \\
4 & \checkmark & \checkmark &            & \textbf{0.9572} & \red{0.9841} & \textbf{45.6986} \\
5 & \checkmark &            & \checkmark & 0.9549 & 0.9828 & 45.5163 \\
6 & \checkmark & \checkmark & \checkmark & 0.9555 & 0.9833 & \red{45.5542} \\
 \hline
\end{tabular}}
\end{table}

The goal of this ablation study is to measure the impact of the losses during training.
As it can be seen from Figure \ref{fig:losses-comparison} and Table \ref{tab:loss-comparison}, three losses are taken into consideration (NCC, SSIM and MSE losses) and three metrics (NCC, SSIM e PSNR metrics).
The model used for training is Swin2SR.

The first observation is related to rows 1, 2 and 3.
From there, we can observe that the NCC loss performs well, as expected, when evaluated using the corresponding NCC metric (row 1).
However, its performance progressively degrades the SSIM and PSNR metrics (see Figure \ref{fig:losses-comparison}).
This should suggests that the NCC loss may not be the most effective choice for optimizing image quality.
With respect to the SSIM, it is evident from the formula that SSIM is more extensive and robust, since it models a wider range of image properties to identify similarities.
This could explain why NCC loss function degrades SSIM metric (row 1) and not the other way around, namely using only the SSIM as loss and the NCC as metric (row 2).
Similarly, we could deduce that the NCC loss may not be well-suited for optimizing the PSNR metric (row 1), obtaining the worst PSNR.
On the contrary, if we use the PSNR as loss, we get a competitive value for the NCC metric (row 3).
This behavior is probably due to the fact that in NCC (loss and metric) we are using global statistics, while in the case of SSIM (loss and metric), MSE loss and PSNR metric, we are using local statistics, i.e. in small sections of the image and single pixel.

Despite these observations, NCC loss still represents an important piece for network training.
Specifically, in Table \ref{tab:loss-comparison} we can observe that, when used in conjunction with the SSIM or MSE loss functions (row 4 and 5), usually the performance of the network is significantly improved, compared to using SSIM or MSE loss alone (row 2 and 3).
This can be explained by the fact that NCC loss introduces limitations on the type of super-resolved images that are generated, in a different but compatible way with SSIM or MSE losses.
Finally, we decided to employ a combination of NCC and SSIM loss since they provided the best results overall.

\subsection{Positional Encoding}
\label{experiment_positional_encoding}

This ablation study aims to evaluate how positional embeddings behave when used together.
The model used as baseline (row 2 of Table \ref{tab:pos_embed}) is the best model from the previous experiment: Swin2SR model (with log-spaced CPB), and both NCC and SSIM losses on training (row 4 of Table \ref{tab:loss-comparison}.

The results shown in Table \ref{tab:pos_embed} indicate that the optimal approach is to employ both per-head bias positional encoding (RPE or log CPB) and per-channel positional encoding (LePE) simultaneously, rather than using either one individually (row 5 vs 1,2 and 3).
Furthermore, using two per-head biases (RPE and log CPB) does not bring substantial differences in performance (see row 1 and 2 vs 6).
Finally, using all three together did not increased substantially the performance to justify the use of all of them (see row 7 vs 4, 5 and 6).

\begin{figure}[ht!]
\centerline{\includegraphics[width=.9\linewidth]{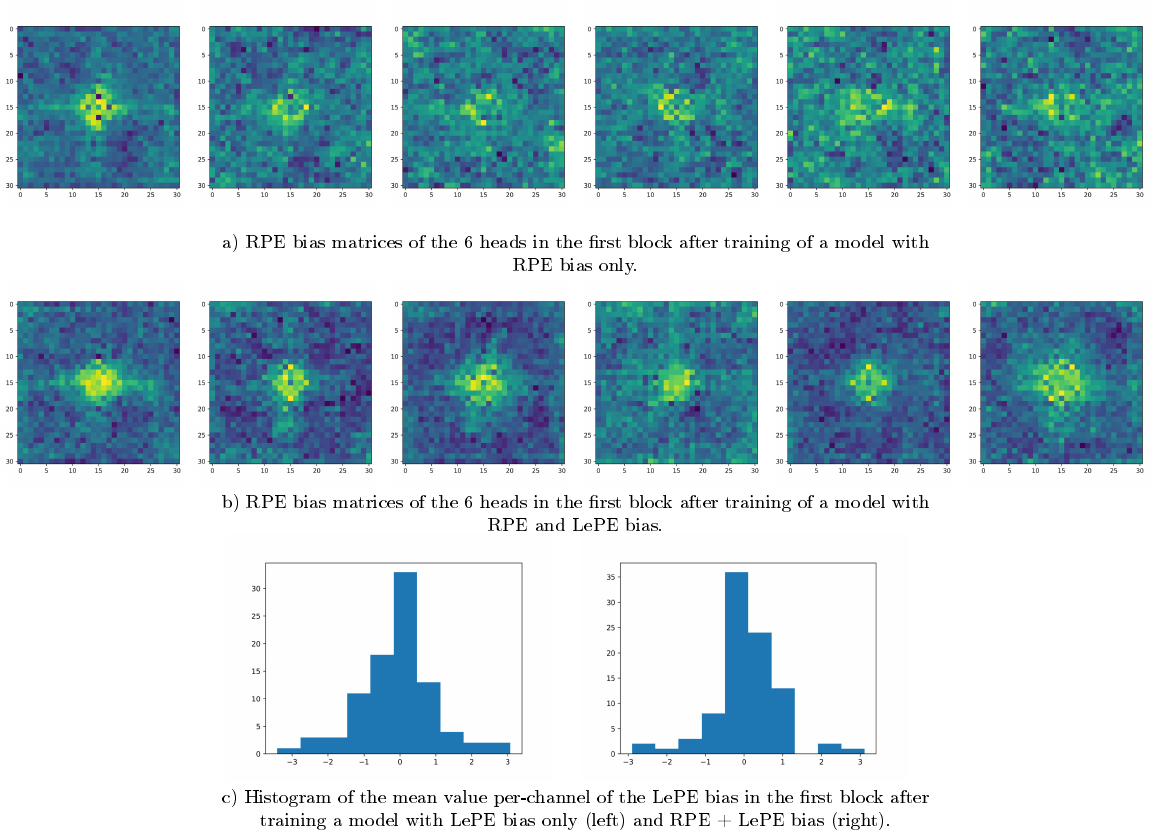}}
    \caption{Visualization of the learnt RPE and LePE bias matrices.}
    \label{fig:pos-enc-visualize}
\end{figure}

Additionally, to demonstrate the validity of our choice, a more detailed analysis of models with different combinations of positional encoding techniques is proposed in Figure~\protect{\ref{fig:pos-enc-visualize}}.
More in detail, the figure presents RPE ($\hat{B} \in \mathbb{R}^{(2M-1)\times(2M-1)}$, where $B$ is the batch size and $M^2$ is the number of patches in a window) and LePE (with shape $B \times C \times 256$, where $C$ is the number of channels) bias matrices after training.
In particular, subfigures (a) and (b) show the matrices of the 6 heads in the first block using the RPE bias without (a) and in combination with (b) LePE bias.
Indeed, the behaviour of RPE is not negatively affected by LePE. On the contrary, the patterns appear consolidated.
Finally, Figure~\protect{\ref{fig:pos-enc-visualize}} (c) presents the histograms of the mean value per-channel of the LePE bias in the first block without RPE (left) and with RPE (right) bias.
Indeed, the distribution of values using LePE alone or in combination with RPE remain for the most part unchanged indicating that the behavior of the LePE is not negatively influenced by the presence of the RPE.

\begin{table}[ht!]
\caption{Ablation study on positional encoding. The best performance is shown in \textbf{bold} and the second-highest performance is shown in \textcolor{red}{red}.\label{tab:pos_embed}}
\centerline{\begin{tabular}{|r|c|c|c|l|l|}
\hline
  & \multicolumn{3}{c|}{Positional Encoding} & \multicolumn{2}{c|}{Performance} \\
  & RPE        & log CPB    & LePE       & SSIM $\uparrow$  & PSNR $\uparrow$ \\ \hline
1          & \checkmark &            &            & 0.9841          & 45.5855 \\
2          &            & \checkmark &            & 0.9841          & 45.6986 \\
3          &            &            & \checkmark & 0.9843          & 45.7278 \\
4          &            & \checkmark & \checkmark & 0.9845          & 45.8046 \\
5          & \checkmark &            & \checkmark & \textbf{0.9847} & \textbf{45.8539} \\
6          & \checkmark & \checkmark &            & 0.9843          & 45.6945 \\
7          & \checkmark & \checkmark & \checkmark & \red{0.9846}    & \red{45.8185} \\
 \hline
\end{tabular}}
\end{table}

\subsection{Mixture of Experts}
\label{experiment_moe}



\begin{table}[ht!]
\caption{Ablation study on Transformer MLP and MoE layer architectures. APC / SPC: active and sparse parameters count inside a MLP and MoE layers. Latency: average amount of seconds required to elaborate a batch of 8 images on a NVIDIA Quadro RTX6000. SM: Smart Merger. Model params: amount of sparse parameters of the full model architecture. Model FLOPs: floating point operations in a forward pass of the full model. The best performance is shown in \textbf{bold} and the second-highest performance is shown in \textcolor{red}{red}}.\label{tab:moe-sm-results}
\centerline{\begin{tabular}{|r|l|c|c|c|l|l|c|r|r|}
\hline
  &           &            & \multicolumn{2}{c|}{MLP \# params} & \multicolumn{2}{c|}{Performance} & Latency & Model & Model \\
  & Arch      & SM         & APC    & SPC      & SSIM $\uparrow$  & PSNR  $\uparrow$  & (s) & params & FLOPs \\ \hline
1 & MLP       &            & 32'670 &  32'670 & \red{0.9847}    & 45.8539          & 0.194  &       2.20M   &   135.54G\\
2 & MoE 8/2   &            & 33'480 & 131'760 & 0.9845          & \red{45.8647}    & 0.223  &       4.59M   &   135.68G\\
3 & MoE 8/2   & \checkmark & 33'499 & 131'779 & \textbf{0.9849} & \textbf{45.9272} & 0.229  &       4.59M   &   138.30G\\
\hline
4 & MoE 8/1   & \checkmark & 34'120 & 262'810 & 0.9837 & 45.4847 & 0.230 & 7.72M & 136.96G\\
5 & MoE 6/2   & \checkmark & 33'319 & 98'839 & 0.9844 & 45.8328 & 0.230 & 3.80M & 138.30G \\
6 & MoE 8/2   & \checkmark & 33'499 & 131'779 & \textbf{0.9849} & \textbf{45.9272} & 0.229  &       4.59M   &   138.30G\\
7 & MoE 10/2  & \checkmark & 33'679 & 164'719 & \red{0.9846} & \red{45.8913} & 0.230 & 5.39M & 138.30G \\
 \hline
\end{tabular}}
\end{table}

The aim of this last ablation study is to evaluate the ability of MoE-SM to rival against the MLP (row 3 vs 1 of Table \ref{tab:moe-sm-results}).
To choose the active and sparse number of experts, we followed the choice made by the authors of Mixtral \cite{jiang2024mixtral} of 8 experts in total but only 2 of them active for each example in input.
To be as fair as possible in the comparison with MLP, each expert has half the number of parameters of an MLP.
In that way, the number of active weights involved in each computation is almost the same.
The 8/2 choice is subsequently validated by a further ablation where the total number of experts available and the number of experts actually used during the forward pass were varied.

Table \ref{tab:moe-sm-results} shows the comparative results.
Baseline model (row 1) is the best model from the previous experiment (row 5 on Table \ref{tab:pos_embed}), which uses MLP.
Regarding our MoE without SM (row 2), the SSIM and PSNR have some fluctuation compared to MLP version (row 1).
Instead, our MoE with SM (row 3) surpasses MLP by $+0.0002$ for SSIM and by $+0.0735$ of PSNR.
On the negative side, the replacement of MLP with MoE slightly increases the average latency, due to the non-optimized sequential execution of the two experts on a single GPU.

With regard to the second ablation study (from row 4 to row 7), reducing the number of active experts (row 4 vs.~row 6) decreases the performance, highlighting the important role played by our SM module.
Also decreasing (row 5 vs.~row 6) or increasing (row 7 vs.~row 6) the number of total experts available does not help.
In our hypothesis, the choice of the number of total experts could also be linked to the amount of data available during training.
For this reason, the strategy of further increasing the number of total experts could be fruitful if the dataset were larger.

\subsection{SISR Experiments}

\noindent\textbf{Quantitative comparison.}
We provide a performance comparison of Swin-MoSE with state-of-the-art SR methods on three upscaling factors.
For $2\times$ upscaling, we used $\text{Sen2Ven}\mu\text{s}$ dataset:
B2, B3, B4, B8 sentinel2 and B3, B4, B7, B11 $\text{Ven}\mu\text{s}$ bands, 10~m and 5~m resolution, respectively.
For $3\times$ upscaling, we used OLI2MSI dataset:
blue, green, red from Landsat-8 and B2, B3, B4 from Sentinel2 bands, 30~m and 10~m resolution, respectively.
For $4\times$ upscaling, we used $\text{Sen2Ven}\mu\text{s}$ dataset:
B5, B6, B7, B8A sentinel2 and B8, B9, B10, B11 $\text{Ven}\mu\text{s}$ bands, 20~m and 5~m resolution, respectively.
For others method, we supposed to use the usual MSE loss for training, compared to us, who use SSIM and NCC together as described in our theory section.


\begin{table}[ht!]
\caption{Quantitative comparison with SOTA models on $\text{Sen2Ven}\mu\text{s}$ and OLI2MSI datasets. GiB, IT/s: gpu memory usage and iterations per second on inference. APC / SPC: active and sparse parameters count inside a MLP and MoE layers. FLOPs: floating point operations in a forward pass of the model. The best performance is shown in \textbf{bold} and the second-highest performance is shown in \textcolor{red}{red}.}
\centerline{\begin{tabular}{|l|l|l|r|r|r|c|c|c|}
\hline
 Model       & \multicolumn{7}{c|}{$\text{Sen2Ve}\mu\text{s}$ $2\times$} \\
     &                 &                  &     &       & \multicolumn{2}{c|}{Params} & \\
    & SSIM $\uparrow$  & PSNR  $\uparrow$ & GiB & IT/s  & APC & SPC & FLOPs \\ \hline
Bicubic                         & 0.9883 & 45.5588 & & & & & \\
SwinIR \cite{liang2021swinir}   & 0.9938 & 48.7064 & 2.37 & 1.32 & 2.10M & 2.10M & 515.90G \\
Swinfir \cite{zhang2022swinfir} & 0.9940 & 48.8532 & 3.12 & 1.28 & 2.52M & 2.52M & 622.42G  \\
Swin2SR \cite{conde2022swin2sr} & 0.9942 & 49.0467 & 3.12 & 1.18 & 2.11M & 2.11M & 526.78G  \\
DAT \cite{conf/iccv/0014ZGKY023}         & \textcolor{red}{0.9946} & \textcolor{red}{49.4699} & 1.46 & 1.95 & 2.14M & 2.14M & 547.22G  \\
\hline
Swin2-MoSE (ours)              & \textbf{0.9948} & \textbf{49.4883} & 2.41 & 1.16 & 2.20M & 4.56M & 543.00G \\
\hline
        & \multicolumn{7}{c|}{$\text{OLI2MSI}$ $3\times$} \\
     &                 &                  &     &       & \multicolumn{2}{c|}{Params} & \\
    & SSIM $\uparrow$  & PSNR  $\uparrow$ & GiB & IT/s  & APC & SPC & FLOPs \\ \hline
Bicubic                         & 0.9768 & 42.1835 & & & & & \\
SwinIR \cite{liang2021swinir}   & 0.9863 & 44.4806 & 3.55 & 0.77 & 2.11M & 2.11M & 809.42G \\
Swinfir \cite{zhang2022swinfir} & 0.9863 & 44.4829 & 4.73 & 0.76 & 2.53M & 2.53M & 975.84G  \\
Swin2SR \cite{conde2022swin2sr} & 0.9881 & 44.9614 & 4.80 & 0.70 & 2.12M & 2.12M & 826.30G  \\
DAT \cite{conf/iccv/0014ZGKY023}         & \textcolor{red}{0.9886} & \textcolor{red}{45.0404} & 2.10 & 1.10 & 2.15M & 2.15M & 858.35G \\
\hline
Swin2-MoSE (ours)              & \textbf{0.9912} & \textbf{45.9414} & 4.88 & 0.52 & 2.20M & 4.56M & 851.76G \\
\hline
        & \multicolumn{7}{c|}{$\text{Sen2Ve}\mu\text{s}$ $4\times$} \\
     &                 &                  &     &       & \multicolumn{2}{c|}{Params} & \\
    & SSIM $\uparrow$  & PSNR  $\uparrow$ & GiB & IT/s  & APC & SPC & FLOPs \\ \hline
Bicubic                         & 0.9674 & 42.0499 & & & & & \\
SwinIR \cite{liang2021swinir}   & 0.9825 & 45.3460 & 0.84 & 5.37 & 2.14M & 2.14M & 131.53G \\
Swinfir \cite{zhang2022swinfir} & 0.9830 & 45.5500 & 0.84 & 5.20 & 2.56M & 2.56M & 158.17G  \\
Swin2SR \cite{conde2022swin2sr} & 0.9828 & 45.4759 & 1.03 & 4.74 & 2.15M & 2.15M & 134.40G  \\
DAT \cite{conf/iccv/0014ZGKY023}         & \textcolor{red}{0.9837} & \textcolor{red}{45.8053} & 0.69 & 6.73 & 2.18M & 2.18M & 139.36G \\
\hline
Swin2-MoSE (ours)              & \textbf{0.9849} & \textbf{45.9272} & 0.86 & 3.91 & 2.23M & 4.59M & 138.30G \\
\hline
\end{tabular}}
\label{tab:sota-results}
\end{table}

The quantitative results shown on Table~\ref{tab:sota-results} demonstrate the superiority of our Swin2-MoSE compared to competing SOTA models.
It improves PSNR by up to $0.377 \sim 0.958$ dB compared to compared with models descended from Swin Transformer like ours.
Compared to SOTA, it slightly outperforms for the 2x task, but maintains a good gap compared to 3x and 4x tasks.
This demonstrates the goodness of our model which behaves significantly better in more complex situations. Additionally, the table presents an evaluation of the computational complexity of our model.
Even if our model has an higher complexity in terms of number of sparse parameters due to the MoE, the number of active parameters and floating-point operations (FLOPs) is still comparable to the other state-of-the-art models.

\begin{figure}[ht!]
\centerline{\includegraphics[width=.9\linewidth]{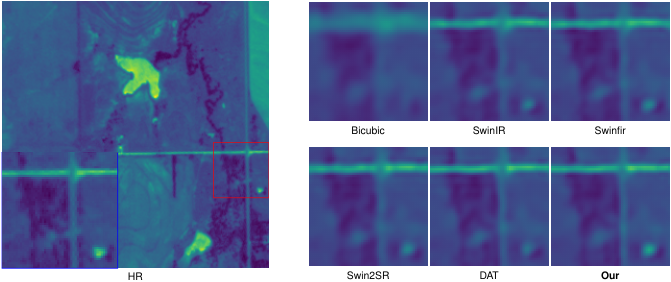}}
\caption{Comparison between 4x SISR methods on $\text{Sen2Ven}\mu\text{s}$ dataset. HR reference is Ven$\mu\text{s}$ B8 (red edge, 5~m). The crop of the HR reference image is shown in the red square, and the zoomed cropped part that was chosen is shown in the blue square.}
\label{fig:sr-comparison1}
\end{figure}

\begin{figure}[ht!]
\centerline{\includegraphics[width=.9\linewidth]{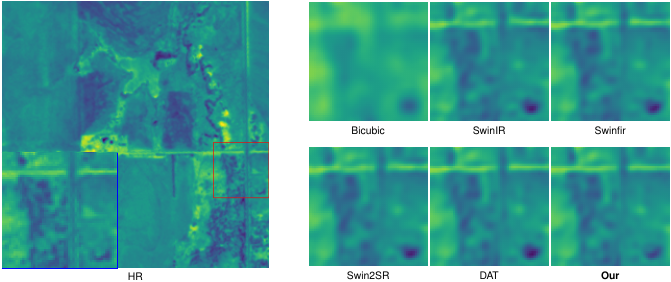}}
\caption{Comparison between 4x SISR methods on $\text{Sen2Ven}\mu\text{s}$ dataset. HR reference is Ven$\mu\text{s}$ B9 (red edge, 5~m). The crop of the HR reference image is shown in the red square, and the zoomed cropped part that was chosen is shown in the blue square.}
\label{fig:sr-comparison2}
\end{figure}

\begin{figure}[ht!]
\centerline{\includegraphics[width=.9\linewidth]{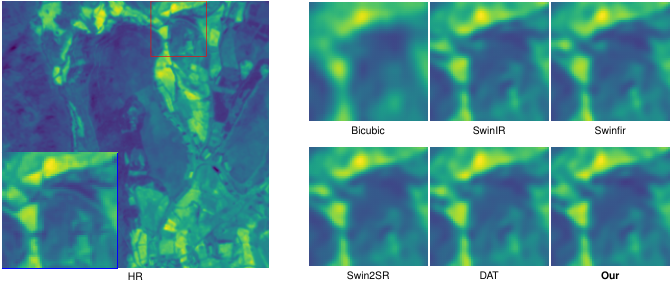}}
\caption{Comparison between 4x SISR methods on $\text{Sen2Ven}\mu\text{s}$ dataset. HR reference is Ven$\mu\text{s}$ B10 (red edge, 5~m). The crop of the HR reference image is shown in the red square, and the zoomed cropped part that was chosen is shown in the blue square.}
\label{fig:sr-comparison3}
\end{figure}

\begin{figure}[ht!]
\centerline{\includegraphics[width=.9\linewidth]{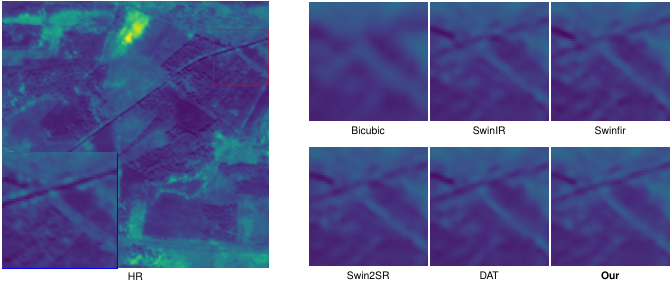}}
\caption{Comparison between 4x SISR methods on $\text{Sen2Ven}\mu\text{s}$ dataset. HR reference is Ven$\mu\text{s}$ B11 (NIR, 5~m). The crop of the HR reference image is shown in the red square, and the zoomed cropped part that was chosen is shown in the blue square.}
\label{fig:sr-comparison4}
\end{figure}

\begin{figure}[ht!]
\centerline{\includegraphics[width=.9\linewidth]{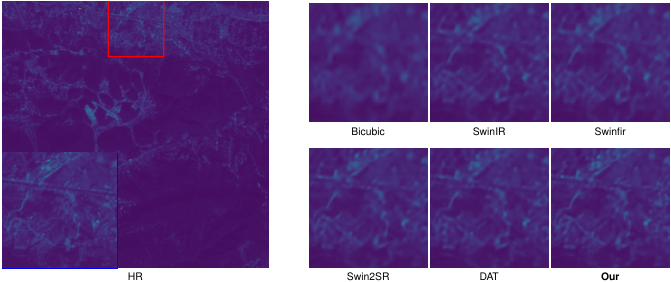}}
\caption{Comparison between 3x SISR methods on OLI2MSI dataset. HR reference is Sentinel-2 B2 (blue, 10~m). The crop of the HR reference image is shown in the red square, and the zoomed cropped part that was chosen is shown in the blue square.}
\label{fig:sr-comparison1-3x}
\end{figure}

\begin{figure}[ht!]
\centerline{\includegraphics[width=.9\linewidth]{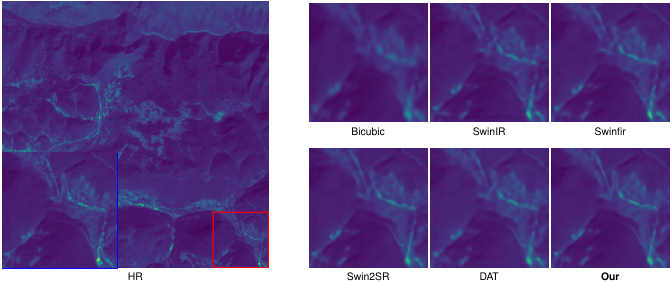}}
\caption{Comparison between 3x SISR methods on OLI2MSI dataset. HR reference is Sentinel-2 B3 (green, 10~m). The crop of the HR reference image is shown in the red square, and the zoomed cropped part that was chosen is shown in the blue square.}
\label{fig:sr-comparison2-3x}
\end{figure}

\begin{figure}[ht!]
\centerline{\includegraphics[width=.9\linewidth]{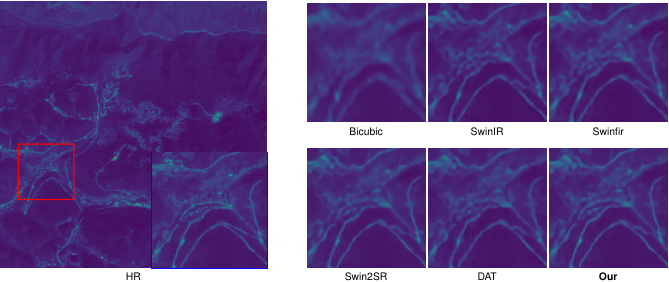}}
\caption{Comparison between 3x SISR methods on OLI2MSI dataset. HR reference is Sentinel-2 B4 (red, 10~m). The crop of the HR reference image is shown in the red square, and the zoomed cropped part that was chosen is shown in the blue square.}
\label{fig:sr-comparison3-3x}
\end{figure}

\noindent\textbf{Qualitative comparison.}
Figures~\ref{fig:sr-comparison1}, \ref{fig:sr-comparison2}, \ref{fig:sr-comparison3}, and \ref{fig:sr-comparison4} illustrate the qualitative comparison of the considered super-resolution (SR) methods and the proposed approach in this study, for the task of $4\times$ SISR on $\text{Sen2Ven}\mu\text{s}$ dataset.
Figures~\ref{fig:sr-comparison1-3x}, \ref{fig:sr-comparison2-3x} and \ref{fig:sr-comparison3-3x} illustrate the qualitative comparison for the task of $3\times$ SISR on OLI2MSI dataset.
The red boxes overlaid on the figures highlight the reconstructed regions that are being compared.
From the images we can observe the better quality and less distorted images generated from our model compared to SOTA models.



\subsection{Semantic Segmentation Experiments}\label{sec4.5}

Finally, we want to demonstrate the effectiveness of the proposed model trained for image super-resolution in improving the performance of a different task, namely semantic segmentation.
In order to simplify the procedure, we take into account as input only the 10~m bands of Sentinel-2: B2, B3, B4 and B8.
In this way, we can use our Swin2-MoSE, previously trained on $2\times$ SISR task on $\text{Sen2Ven}\mu\text{s}$ dataset and frozen, as feature extractor for the semantic segmentation model, Foreground-Aware Relation Network (FarSeg) model \cite{zheng2020foreground} in this case.
Our goal is to prove that our model is able to provide enriched features the segmentation model, increasing its classification capabilities.

More in details, the tested training configurations are the following:
1) The baseline, called FarSeg, employs the FarSeg model as it is, training it with the four-channels SeasoNet satellite images as input.
2) The FarSeg-S2MFE (FarSeg with the Swin2-MoSE Feature Extractor) is using our pretrained and frozen Swin2-MoSE model, without the HQ Image Reconstruction module, as feature extractor (see Feature Extractor in Figure \ref{swin2-moe-arch}).
In this case, the satellite images are firstly passed through our feature extractor and then, the output features (with a shape $T\times128\times128$, where $T=90$ represents the embedding dimension) are forwarded as input to the FarSeg model.
3) To be as fair as possible with the baseline, we prepared another baseline, called FarSeg++, where the number of channels of the input images ($4\times120\times120$) are firstly increased to T with a 2D convolution (kernel size 1, padding 0), to obtain the shape $T\times128\times128$ (the same of the output of our feature extractor), and then are passed as input to FarSeg model.
It is worth noting that, compared to the previously defined configurations, in this last case there is one more convolution that is trained.

\begin{figure}[ht!]
    \centering
    \begin{subfigure}{0.49\linewidth}
        \includegraphics[width=1.0\linewidth]{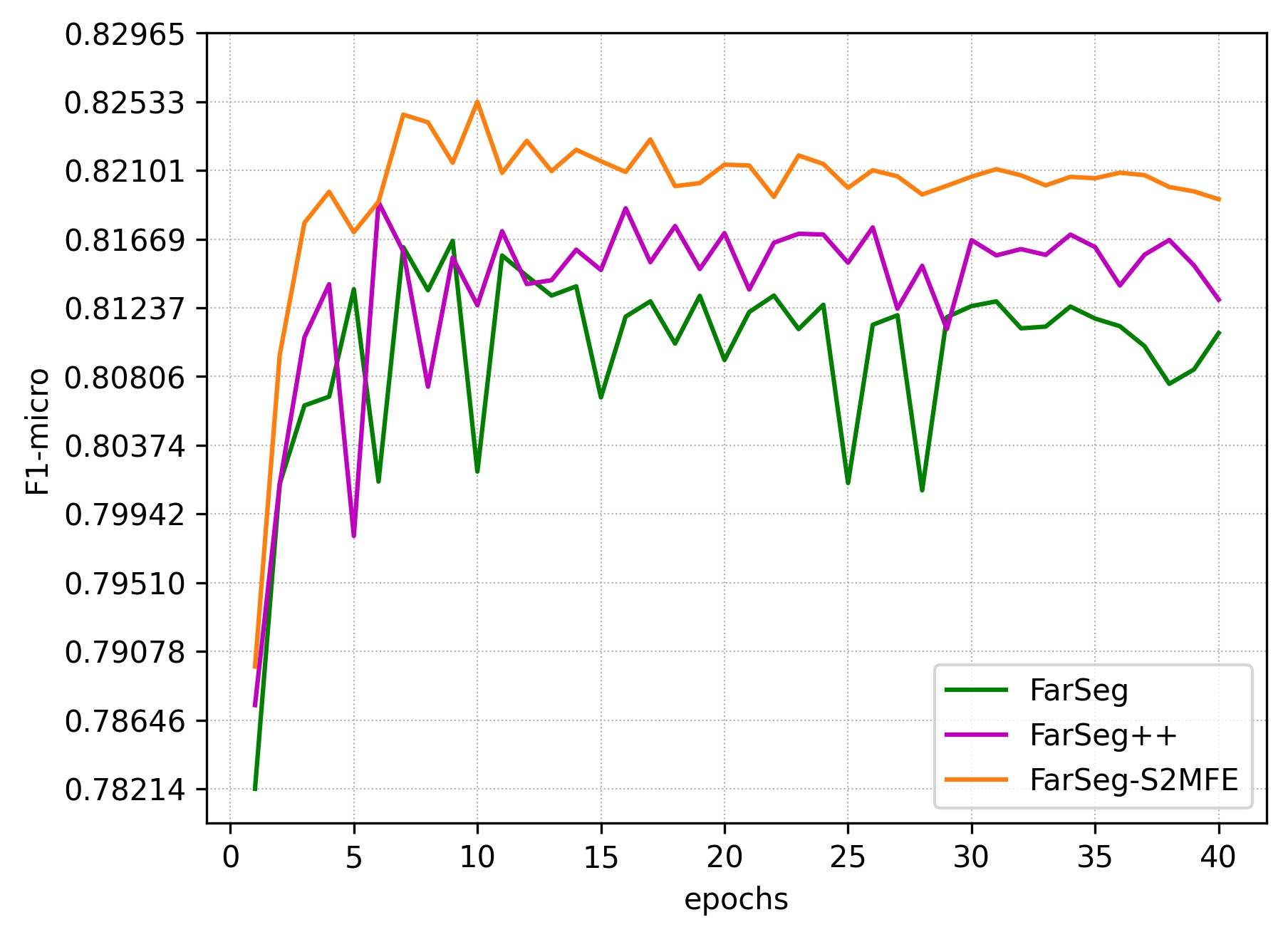}
        \label{fig:ssegm-f1-micro}
    \end{subfigure}
    \begin{subfigure}{0.49\linewidth}
        \includegraphics[width=1.0\linewidth]{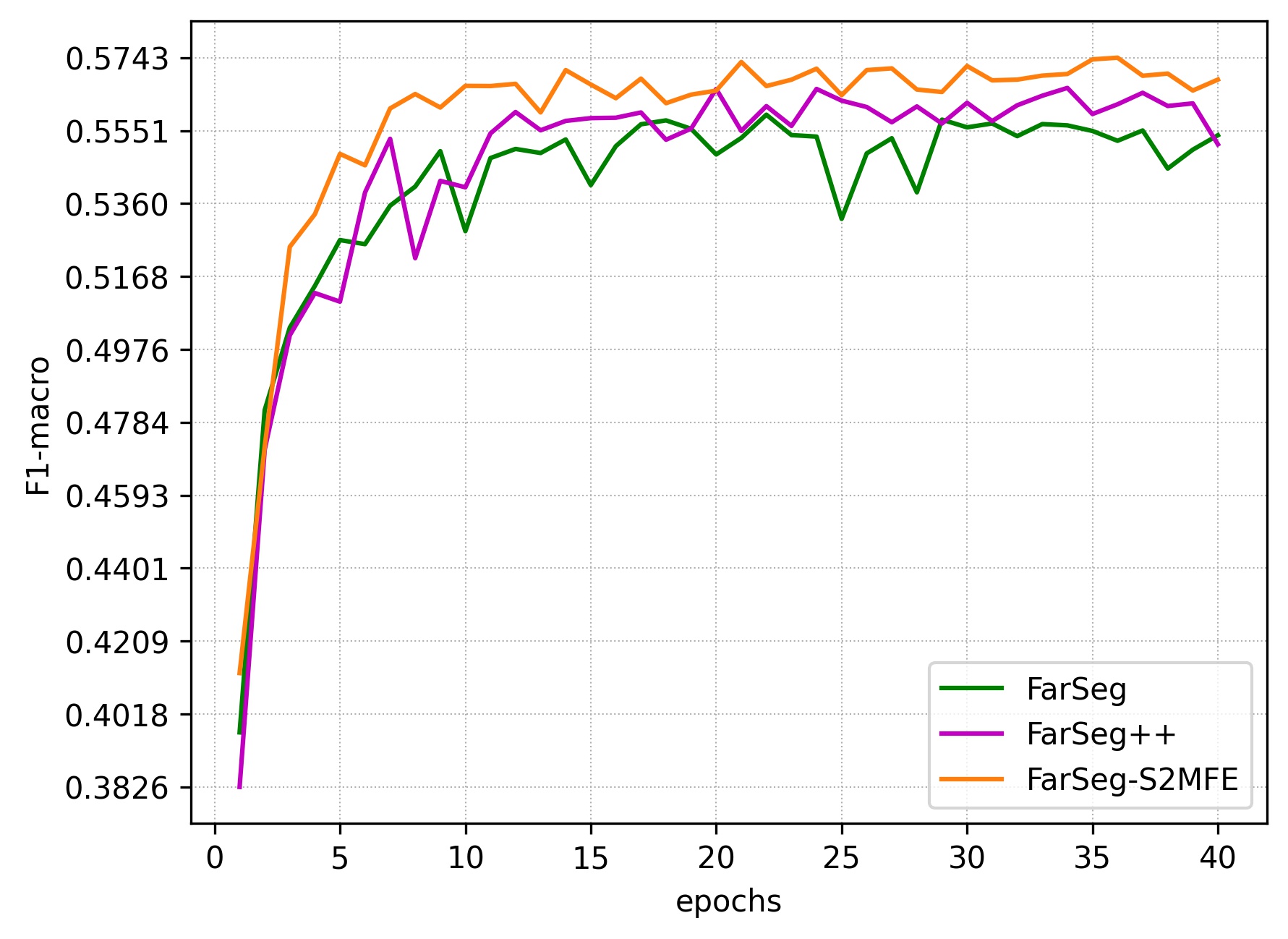}
        \label{fig:ssegm-f1-macro}
    \end{subfigure}
 \caption{Results for the Semantic Segmentation task on SeasoNet dataset. F1-micro and F1-macro are the metrics employed.}
\label{fig:ssegm_results}
\end{figure}

As shown in Figure \ref{fig:ssegm_results}, thanks to our model, FarSeg-S2MFE outperforms both baselines for both metrics F1-macro and F1-micro during all training, demonstrating that it can generate meaningful features for the downstream task of semantic segmentation.
These findings suggest that the use of our super-resolution technique significantly improves the performance of the trained semantic segmentation model.

 \section{Conclusions}\label{conclusions}

In this work, we presented Swin V2 Mixture of Super-resolution Experts (Swin2-MoSE), a novel End-to-End Single-Image Super-Resolution model tailored on remote sensing that introduced several novelties w.r.t. the state of the art.
Firstly, we introduced an an enhanced Sparsely-Gated Mixture-of-Experts (MoE-SM) inside the Transformer architecture blocks
increasing the model performance.
MoE-SM was designed with a new advanced merging layer (Smart Merger, SM) in order to merge the output of individual experts, and with a new way to split the work between experts, defining a new per-example strategy instead of the commonly used per-token one.
Secondly, after an analysis of how positional encodings interact with each other, a per-channel bias (Locally-enhanced Positional Encoding, LePE \cite{dong2022cswin}) and a per-head bias (Relative Position Encoding, RPE \cite{liu2021swin}) were integrated in the architecture.
Thirdly, we proposed a combination of Normalized-Cross-Correlation (NCC) and Structural Similarity Index Measure (SSIM) losses for training, in order to avoid typical limitations of the Mean Square Error (MSE) loss.
Finally, several experiments performed on multiple super-resolution satellite images datasets proved that our system is able to surpass the current SOTA both in SSIM and PSNR metrics.
Our Swin2-MoSE outperforms competitor Swin-based models by up to $0.377 \sim 0.958$ dB (PSNR) and $0.0006 \sim 0.0031$ (SSIM) on task of upscaling resolution of factor $2\times$ ($\text{Sen2Ven}\mu\text{s}$ dataset), $3\times$ (OLI2MSI dataset) and $4\times$ ($\text{Sen2Ven}\mu\text{s}$ dataset), respectively.
Compared to SOTA, it achieves better overall performance, especially for more difficult tasks such as 3x and 4x, where it shows great potential.
Furthermore, an additional experiment on semantic segmentation showed the capability of our model to enhance the performance of different tasks other than SR alone.
\section*{Acknowledgments}

Project ECS\_00000033\_ECOSISTER funded under the National Recovery and Resilience Plan (NRRP), Mission 4 Component 2 Investment 1.5 - funded by the European Union – NextGenerationEU.\\
This research benefits from the HPC (High Performance Computing) facility of the University of Parma, Italy.

\bibliography{main_clean}

\begin{thebibliography}{10}
\providecommand \doibase [0]{http://dx.doi.org/}%

\bibitem{chen2022real}
Chen H, He X, Qing L, et al. Real-world single image super-resolution: A brief
  review. {\it Information Fusion.} 2022\string;79\string:124--145.

\bibitem{chauhan2023deep}
Chauhan K, Patel S, Kumhar M, et al. Deep Learning-based Single-image
  Super-resolution: A comprehensive review. {\it IEEE Access.} 2023.

\bibitem{al2024single}
Al-Mekhlafi H, Liu S. Single image super-resolution: a comprehensive review and
  recent insight. {\it Frontiers of Computer Science.}
  2024\string;18(1)\string:181702.

\bibitem{liu2021research}
Liu H, Qian Y, Zhong X, Chen L, Yang G. Research on super-resolution
  reconstruction of remote sensing images: A comprehensive review. {\it Optical
  Engineering.} 2021\string;60(10)\string:100901--100901.

\bibitem{fernandez2017single}
Fernandez-Beltran R, Latorre-Carmona P, Pla F. Single-frame super-resolution in
  remote sensing: A practical overview. {\it International journal of remote
  sensing.} 2017\string;38(1)\string:314--354.

\bibitem{wang2022review}
Wang X, Yi J, Guo J, et al. A review of image super-resolution approaches based
  on deep learning and applications in remote sensing. {\it Remote Sensing.}
  2022\string;14(21)\string:5423.

\bibitem{wang2022comprehensive}
Wang P, Bayram B, Sertel E. A comprehensive review on deep learning based
  remote sensing image super-resolution methods. {\it Earth-Science Reviews.}
  2022\string:104110.

\bibitem{kossmann2022seasonet}
Ko{\ss}mann D, Brack V, Wilhelm T. Seasonet: A seasonal scene classification,
  segmentation and retrieval dataset for satellite imagery over germany. In:
  Proceedings of the IGARSS 2022-2022 IEEE International Geoscience and Remote
  Sensing Symposium.  2022\string:243--246.

\bibitem{reiersen2022reforestree}
Reiersen G, Dao D, L{\"u}tjens B, et al. ReforesTree: A dataset for estimating
  tropical forest carbon stock with deep learning and aerial imagery. In:
  Proceedings of the AAAI Conference on Artificial Intelligence.
  2022\string:12119--12125.

\bibitem{daudt2018urban}
Daudt RC, Le~Saux B, Boulch A, Gousseau Y. Urban change detection for
  multispectral earth observation using convolutional neural networks. In:
  Proceedings of the IEEE International Geoscience and Remote Sensing Symposium
  IGARSS.  2018\string:2115--2118.

\bibitem{anger2020fast}
Anger J, Ehret T, Franchis dC, Facciolo G. Fast and accurate multi-frame
  super-resolution of satellite images. {\it ISPRS Annals of the
  Photogrammetry, Remote Sensing and Spatial Information Sciences.}
  2020\string;1\string:57--64.

\bibitem{lafenetre2023handheld}
Lafenetre J, Nguyen NL, Facciolo G, Eboli T. Handheld burst super-resolution
  meets multi-exposure satellite imagery. In: Proceedings of the IEEE/CVF
  Conference on Computer Vision and Pattern Recognition.
  2023\string:2055--2063.

\bibitem{nguyen2023l1bsr}
Nguyen NL, Anger J, Davy A, Arias P, Facciolo G. L1BSR: Exploiting Detector
  Overlap for Self-Supervised Single-Image Super-Resolution of Sentinel-2 L1B
  Imagery. In: Proceedings of the IEEE/CVF Conference on Computer Vision and
  Pattern Recognition.  2023\string:2012--2022.

\bibitem{conde2022swin2sr}
Conde MV, Choi UJ, Burchi M, Timofte R. Swin2SR: Swinv2 transformer for
  compressed image super-resolution and restoration. In: European Conference on
  Computer Vision.  2022\string:669--687.

\bibitem{shazeer2017outrageously}
Shazeer N, Mirhoseini A, Maziarz K, et al. Outrageously large neural networks:
  The sparsely-gated mixture-of-experts layer. {\it arXiv preprint
  arXiv:1701.06538.} 2017.

\bibitem{dong2022cswin}
Dong X, Bao J, Chen D, et al. Cswin transformer: A general vision transformer
  backbone with cross-shaped windows. In: Proceedings of the IEEE/CVF
  Conference on Computer Vision and Pattern Recognition.
  2022\string:12124--12134.

\bibitem{liu2021swin}
Liu Z, Lin Y, Cao Y, et al. Swin transformer: Hierarchical vision transformer
  using shifted windows. In: Proceedings of the IEEE/CVF international
  conference on computer vision.  2021\string:10012--10022.

\bibitem{dong2015image}
Dong C, Loy CC, He K, Tang X. Image super-resolution using deep convolutional
  networks. {\it IEEE transactions on pattern analysis and machine
  intelligence.} 2015\string;38(2)\string:295--307.

\bibitem{dai2019second}
Dai T, Cai J, Zhang Y, Xia ST, Zhang L. Second-order attention network for
  single image super-resolution. In: Proceedings of the IEEE/CVF conference on
  computer vision and pattern recognition.  2019\string:11065--11074.

\bibitem{kim2016accurate}
Kim J, Lee JK, Lee KM. Accurate image super-resolution using very deep
  convolutional networks. In: Proceedings of the IEEE conference on computer
  vision and pattern recognition.  2016\string:1646--1654.

\bibitem{ledig2017photo}
Ledig C, Theis L, Husz{\'a}r F, et al. Photo-realistic single image
  super-resolution using a generative adversarial network. In: Proceedings of
  the IEEE conference on computer vision and pattern recognition.
  2017\string:4681--4690.

\bibitem{wang2018esrgan}
Wang X, Yu K, Wu S, et al. Esrgan: Enhanced super-resolution generative
  adversarial networks. In: Proceedings of the European conference on computer
  vision (ECCV) workshops.  2018\string:0--0.

\bibitem{niu2020single}
Niu B, Wen W, Ren W, et al. Single image super-resolution via a holistic
  attention network. In: Proceedsings of Computer Vision -- ECCV 2020: 16th
  European Conference, Glasgow, Part XII 16.  2020\string:191--207.

\bibitem{wang2022uformer}
Wang Z, Cun X, Bao J, Zhou W, Liu J, Li H. Uformer: A general u-shaped
  transformer for image restoration. In: Proceedings of the IEEE/CVF conference
  on computer vision and pattern recognition.  2022\string:17683--17693.

\bibitem{liang2021swinir}
Liang J, Cao J, Sun G, Zhang K, Van~Gool L, Timofte R. Swinir: Image
  restoration using swin transformer. In: Proceedings of the IEEE/CVF
  international conference on computer vision.  2021\string:1833--1844.

\bibitem{zhang2022swinfir}
Zhang D, Huang F, Liu S, Wang X, Jin Z. Swinfir: Revisiting the swinir with
  fast fourier convolution and improved training for image super-resolution.
  {\it arXiv preprint arXiv:2208.11247.} 2022.

\bibitem{liu2022swin}
Liu Z, Hu H, Lin Y, et al. Swin transformer v2: Scaling up capacity and
  resolution. In: Proceedings of the IEEE/CVF conference on computer vision and
  pattern recognition.  2022\string:12009--12019.

\bibitem{riquelme2021scaling}
Riquelme C, Puigcerver J, Mustafa B, et al. Scaling vision with sparse mixture
  of experts. {\it Advances in Neural Information Processing Systems.}
  2021\string;34\string:8583--8595.

\bibitem{jordan1994hierarchical}
Jordan MI, Jacobs RA. Hierarchical mixtures of experts and the EM algorithm.
  {\it Neural computation.} 1994\string;6(2)\string:181--214.

\bibitem{conf/iccv/0014ZGKY023}
Chen Z, Zhang Y, Gu J, Kong L, Yang X, Yu F. Dual Aggregation Transformer for
  Image Super-Resolution. In: Proceedings of the Computer Vision -- ICCV.
  2023.

\bibitem{tuna2018srcnnrs}
Caglayan~Tuna GU, Sertel E. Single-frame super resolution of remote-sensing
  images by convolutional neural networks. {\it International Journal of Remote
  Sensing.} 2018\string;39(8)\string:2463-2479.
\newblock \href {\doibase 10.1080/01431161.2018.1425561} {doi:
  10.1080/01431161.2018.1425561}

\bibitem{dong2016vdsrrs}
Dong C, Loy CC, He K, Tang X. Image Super-Resolution Using Deep Convolutional
  Networks. {\it IEEE Transactions on Pattern Analysis and Machine
  Intelligence.} 2016\string;38(2)\string:295-307.
\newblock \href {\doibase 10.1109/TPAMI.2015.2439281} {doi:
  10.1109/TPAMI.2015.2439281}

\bibitem{karwowska2022review}
Karwowska K, Wierzbicki D. Using Super-Resolution Algorithms for Small
  Satellite Imagery: A Systematic Review. {\it IEEE Journal of Selected Topics
  in Applied Earth Observations and Remote Sensing.}
  2022\string;15\string:3292-3312.
\newblock \href {\doibase 10.1109/JSTARS.2022.3167646} {doi:
  10.1109/JSTARS.2022.3167646}

\bibitem{lei2017lgcnet}
Lei S, Shi Z, Zou Z. Super-Resolution for Remote Sensing Images via
  Local–Global Combined Network. {\it IEEE Geoscience and Remote Sensing
  Letters.} 2017\string;14(8)\string:1243-1247.
\newblock \href {\doibase 10.1109/LGRS.2017.2704122} {doi:
  10.1109/LGRS.2017.2704122}

\bibitem{xu2018dmcn}
Xu W, XU G, Wang Y, Sun X, Lin D, WU Y. High Quality Remote Sensing Image
  Super-Resolution Using Deep Memory Connected Network. In: Proceedings of the
  IEEE International Geoscience and Remote Sensing Symposium IGARSS.
  2018\string:8889-8892

\bibitem{ren2021ercnn}
Ren C, He X, Qing L, Wu Y, Pu Y. Remote sensing image recovery via enhanced
  residual learning and dual-luminance scheme. {\it Knowledge-Based Systems.}
  2021\string;222\string:107013.
\newblock \href {\doibase https://doi.org/10.1016/j.knosys.2021.107013} {doi:
  https://doi.org/10.1016/j.knosys.2021.107013}

\bibitem{guo2022ndsrgan}
Guo M, Zhang Z, Liu H, Huang Y. NDSRGAN: A Novel Dense Generative Adversarial
  Network for Real Aerial Imagery Super-Resolution Reconstruction. {\it Remote
  Sensing.} 2022\string;14(7).
\newblock \href {\doibase 10.3390/rs14071574} {doi: 10.3390/rs14071574}

\bibitem{zhang2022rbanunet}
Zhang J, Xu T, Li J, Jiang S, Zhang Y. Single-Image Super Resolution of Remote
  Sensing Images with Real-World Degradation Modeling. {\it Remote Sensing.}
  2022\string;14(12).
\newblock \href {\doibase 10.3390/rs14122895} {doi: 10.3390/rs14122895}

\bibitem{xiao2023transformerRS}
Yi~Xiao JH, Zhang L. Remote sensing image super-resolution via cross-scale
  hierarchical transformer. {\it Geo-spatial Information Science.}
  2023\string;0(0)\string:1-17.
\newblock \href {\doibase 10.1080/10095020.2023.2288179} {doi:
  10.1080/10095020.2023.2288179}

\bibitem{tu2022swcgan}
Tu J, Mei G, Ma Z, Piccialli F. SWCGAN: Generative Adversarial Network
  Combining Swin Transformer and CNN for Remote Sensing Image Super-Resolution.
  {\it IEEE Journal of Selected Topics in Applied Earth Observations and Remote
  Sensing.} 2022\string;15\string:5662-5673.
\newblock \href {\doibase 10.1109/JSTARS.2022.3190322} {doi:
  10.1109/JSTARS.2022.3190322}

\bibitem{vaswani2017attention}
Vaswani A, Shazeer N, Parmar N, et al. Attention is all you need. {\it Advances
  in neural information processing systems.} 2017\string;30.

\bibitem{shaw2018self}
Shaw P, Uszkoreit J, Vaswani A. Self-attention with relative position
  representations. {\it arXiv preprint arXiv:1803.02155.} 2018.

\bibitem{10.1007/978-3-031-19797-0_33}
Liang J, Zeng H, Zhang L. Efficient and Degradation-Adaptive Network
  for Real-World Image Super-Resolution. In:  Avidan S, Brostow G, Ciss{\'e}
  M, Farinella GM, Hassner T. \kern-2pt, eds. {\it Proceedings of the Computer
  Vision -- ECCV}Proceedings of the Computer Vision -- ECCV. Springer Nature
  Switzerland 2022; Cham\string:574--591.

\bibitem{he2024frequency}
He X, Yan K, Li R, Xie C, Zhang J, Zhou M. Frequency-Adaptive Pan-Sharpening
  with Mixture of Experts. {\it arXiv preprint arXiv:2401.02151.} 2024.

\bibitem{hwang2023tutel}
Hwang C, Cui W, Xiong Y, et al. Tutel: Adaptive mixture-of-experts at scale.
  {\it Proceedings of Machine Learning and Systems.} 2023\string;5.

\bibitem{fedus2022switch}
Fedus W, Zoph B, Shazeer N. Switch transformers: Scaling to trillion parameter
  models with simple and efficient sparsity. {\it The Journal of Machine
  Learning Research.} 2022\string;23(1)\string:5232--5270.

\bibitem{jiang2024mixtral}
Jiang AQ, Sablayrolles A, Roux A, et al. Mixtral of experts. {\it arXiv
  preprint arXiv:2401.04088.} 2024.

\bibitem{sajjadi2017enhancenet}
Sajjadi MS, Scholkopf B, Hirsch M. Enhancenet: Single image super-resolution
  through automated texture synthesis. In: Proceedings of the IEEE
  international conference on computer vision.  2017\string:4491--4500.

\bibitem{wang2004image}
Wang Z, Bovik AC, Sheikh HR, Simoncelli EP. Image quality assessment: from
  error visibility to structural similarity. {\it IEEE transactions on image
  processing.} 2004\string;13(4)\string:600--612.

\bibitem{michel2022sen2venmus}
Michel J, Vinasco-Salinas J, Inglada J, Hagolle O. Sen2ven$\mu$s, a dataset for
  the training of sentinel-2 super-resolution algorithms. {\it Data.}
  2022\string;7(7)\string:96.

\bibitem{wang2021multisensor}
Wang J, Gao K, Zhang Z, et al. Multisensor remote sensing imagery
  super-resolution with conditional GAN. {\it Journal of Remote Sensing.} 2021.

\bibitem{zheng2020foreground}
Zheng Z, Zhong Y, Wang J, Ma A. Foreground-aware relation network for
  geospatial object segmentation in high spatial resolution remote sensing
  imagery. In: Proceedings of the IEEE/CVF conference on computer vision and
  pattern recognition.  2020\string:4096--4105.

\end{thebibliography}

\end{document}